\documentclass[letterpaper, 10 pt, conference]{ieeeconf} 
\IEEEoverridecommandlockouts                             

\usepackage{amsmath}
\usepackage{amsfonts}
\usepackage{graphicx}
\usepackage[dvipsnames]{xcolor}
\usepackage{wrapfig}
\usepackage{stfloats}
\usepackage{caption}
\usepackage{booktabs}
\usepackage{longtable}
\usepackage{subfigure}
\usepackage{hyperref}

\DeclareMathOperator*{\argmin}{arg\,min}

\newcommand{\ie}{\textit{i}.\textit{e}. }
\newcommand{\eg}{\textit{e}.\textit{g}. }
\newcommand\sref{Section~\ref}

\newcommand\fref{Fig.~\ref}
\newcommand\tref{Table~\ref}

\newcommand\blfootnote[1]{%
\begingroup
\renewcommand\thefootnote{}\footnote{#1}%
\addtocounter{footnote}{-1}%
\endgroup
}

\newif\ifarxiv
\title{\LARGE \bf
RoboTAP: Tracking Arbitrary Points for Few-Shot Visual Imitation
}
\author{
Mel Vecerik$^{1,2,*}$,
Carl Doersch$^{1,*}$,
Yi Yang$^{1}$,
Todor Davchev$^{1}$,
Yusuf Aytar$^{1}$,
Guangyao Zhou$^{1}$ \\
Raia Hadsell$^{1}$,
Lourdes Agapito$^{2}$,
Jon Scholz$^{1,*}$
}
\begin{document}

\arxivtrue

\twocolumn[{
\renewcommand\twocolumn[1][]{#1}
\maketitle
\begin{center}
    \vspace{-0.3cm}
    \centering
    \captionsetup{type=figure}
    \includegraphics[width=.98\textwidth]{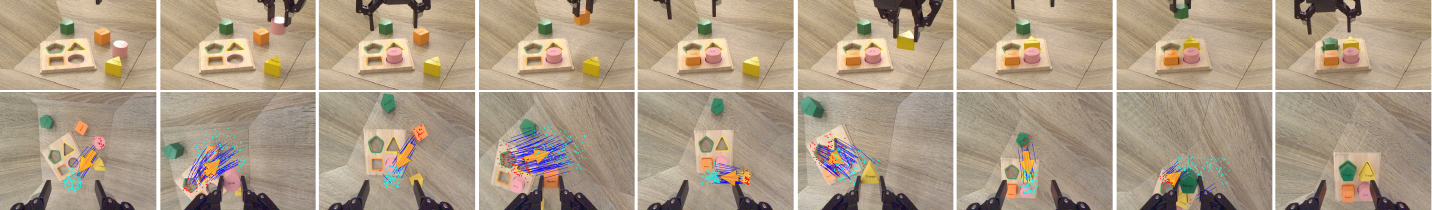}
    \captionof{figure}{
    \small An example of RoboTAP using points automatically selected from few demos ($\leq 6$) to define a long horizon behaviour.
    First row is illustrative, second row is what the agent sees. At every stage, the system identifies the current location of ``active'' points relevant to the stage (red). 
    Given the goal locations for each point, from the demos (cyan), a desired motion for each point is produced (blue lines), and converted to a robot action using a generalized 4D visual-servoing primitive, which operates with arbitrary points.
    }
    \label{fig:highlight}
\end{center}
}]

\thispagestyle{plain}
\pagestyle{plain}
\begin{abstract}
For robots to be useful outside labs and specialized factories we need a way to teach them new useful behaviors quickly.
Current approaches lack either the generality to onboard new tasks without task-specific engineering, or else lack the data-efficiency to do so in an amount of time that enables practical use.
In this work we explore \textit{dense tracking} as a representational vehicle to allow faster and more general learning from demonstration.
Our approach utilizes Track-Any-Point (TAP) models to isolate the relevant motion in a demonstration, and parameterize a low-level controller to reproduce this motion across changes in the scene configuration.
We show this results in robust robot policies that can solve complex object-arrangement tasks such as shape-matching, stacking, and even full path-following tasks such as applying glue and sticking objects together, all from demonstrations that can be collected in minutes.
\blfootnote{$*$Main contacts: {\tt\small \{vec,doersch,jscholz\}@google.com}}%
\blfootnote{$^{1}$Google DeepMind}%
\blfootnote{$^{2}$Department of Computer Science at University College London}%
\blfootnote{Project website: \url{https://robotap.github.io}}%
\end{abstract}

\section{INTRODUCTION}

Imagine if you could track an arbitrary number of points in space, in any scene, through occlusions, motion, and deformation -- how might it simplify the manipulation problem?
Recent advancements in \textit{dense-tracking} \cite{doersch2022tap} have provided exactly this capability, but it has yet to be explored in a manipulation setting.
In this paper we investigate dense-tracking as a perceptual primitive for manipulation, with the goal of producing a single easy-to-use system 
that can solve a wide range of problems without task-specific engineering.

This objective is closely related to recent work on foundation-models for robotics~\cite{saycan2022arxiv,bousmalis2023robocat}, which treats manipulation across embodiments and tasks as a large-scale sequence-modeling problem.
These approaches are incredibly general and powerful, especially when utilizing models pre-trained on large multi-modal datasets~\cite{brohan2022rt,liang2023code}, but they tend to be data-hungry and expensive to train.
From this perspective, our primary interest is in understanding the representational and computational bottlenecks in the manipulation problem, in order to improve the scalability and performance of generalist robotic systems.

Our hypothesis is that much of the complexity in low-level manipulation can be reduced to three fundamental operations: (1) identifying \textbf{what} is relevant in the current frame, (2) identifying \textbf{where} it is, and (3) identifying \textbf{how} to move it in desired directions.
We show that all three operations can be parameterized via dense-tracking, yielding a general formulation for manipulation that does not require task-specific engineering.
Furthermore, we find that point tracks can provide an \textit{interface} between these different operations, allowing us to factorize the problem into simple low-dimensional functions which, together, can express a large space of possible behaviors.

The behaviors of interest span a range of real-world problems involving precise multi-object rearrangement, \eg pick-and-place, (high-clearance) insertion, and stacking.
Our approach, RoboTAP, allows few-shot imitation learning of these behaviors in minutes, which we highlight via a task that involves grasping a glue-stick, applying it to a surface, mating the two parts, and placing the result at a desired location. 
This task requires over 1000 precise real-valued actions to be successful, and is messy and irreversible, which renders it infeasible for  approaches requiring large-scale data-gathering.

For all of these tasks, RoboTAP automatically extracts the individual motions, the relevant points for each motion, goal locations for those points, and a generates a plan that can be tracked by a low-level visual servoing primitive.
While currently less general than fully end-to-end approaches, our approach can be trained on as few as 4-6 demonstrations per task, does not require action-supervision, and effortlessly generalizes across clutter and object pose randomization.
Our main contributions are as follows: 
(1) RoboTAP, a formulation of the multi-task manipulation problem in terms of dense-tracking,
(2) Concrete implementations of RoboTAP's \textit{what}, \textit{where}, and \textit{how} problems in the form of visual-saliency, temporal-alignment, and visual-servoing,
(3) A new dense-tracking dataset with ground-truth human annotations tailored for our tasks and evaluated on the TAP-Vid Benchmark focusing on Real-World Robotics Manipulation, and
(4) Empirical results that characterize the success and failure modes of RoboTAP on a range of manipulation tasks involving precise multi-body rearrangement, deformable objects, and irreversible-actions.
\begin{figure*}[t!]
  \centering
  \includegraphics[width=\textwidth]{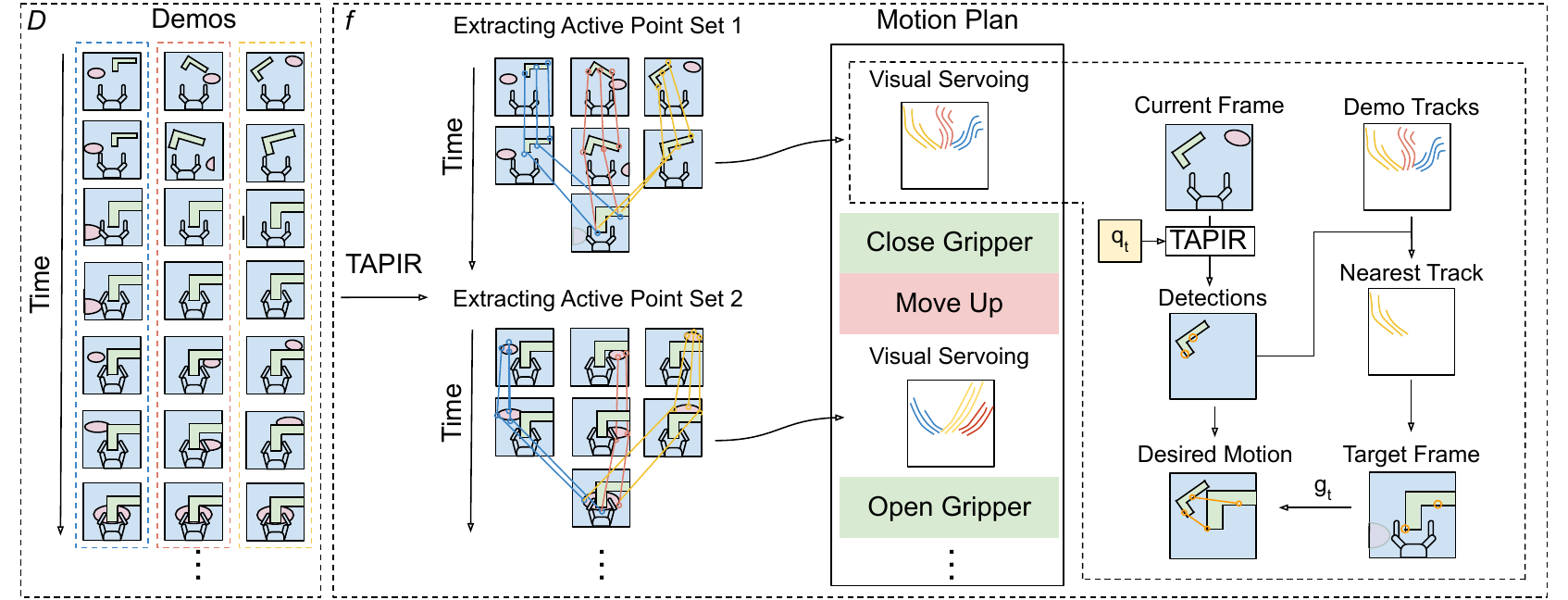}
  \caption{\small 
  Here we describe the core of the RoboTAP approach.
  Given a set of demonstrations D, we first track densely using TAPIR.
  Next, we temporally segment the demonstrations into stages, and then automatically discover the \textit{active point set} $q$ for each stage, which covers the object whose motion is relevant at that stage of the action.
  We then form a motion plan that can be executed on the robot, which consists of stages of servoing to imitate visual motions, and basic motor primitives like closing and opening the gripper.
  Visual servoing is accomplished by detecting points $q$ using TAPIR, finding the nearest demonstration which shows how those points should move, and finding a single nearby frame that can be used as a motion target.
  The displacement between the points in the target frame ($g$) and the online TAPIR detections is used as a motion target for classical visual servoing, yielding surprisingly complex and robust behavior.
  }
  \label{fig:overview}
  \vspace{-0.5cm}
\end{figure*}

\section{Related work}
\label{sec:related_work}
\subsection{Vision-based Manipulation}
\subsubsection{End-to-End Approaches}
Regressing actions directly from images in an end-to-end manner is one established way for utilising visual inputs~\cite{bousmalis2023robocat, kalashnikov2018qt, yen2020learning, lee2020making, luo2021robust}.
While in theory this allows for creation of arbitrarily complex policies in practice this often requires a large amount of data including the target objects or classes.

\subsubsection{Pose-based Approaches}
Another approach is to define a policy on top of object poses regressed from images~\cite{zhu2014single, xiang2017posecnn, deng2020self, chen2022sim, chen2023texpose}.
Pose is a strong signal for object state, but is not well-defined for deformable or symmetric objects, and is difficult and time-consuming to supervise accurately and generally.

\subsubsection{Keypoint-based Approaches}
Within RoboTAP, point locations are detected from an RGB camera and converted to arm motions in a visual feedback-loop, which is commonly referred to as image-based visual servoing~\cite{hill1979real, pomares2019visual}. 
Existing autonomous visual servoing techniques typically rely on hand-designed target detectors or confidence maps~\cite{vahrenkamp2008visual, kragic2002survey, ribeiro2021real}, which lack the generality required for few-shot imitation in arbitrary scenes.

A closely-related approach is to define a small sparse set of keypoints for a given object \textit{class}~\cite{manuelli2019kpam, vecerik2020s3k, das2021model}
Choosing a different point set however requires a retraining of the whole model.
Methods such as TACK~\cite{vecerik2022few} or DON~\cite{florence2018dense} generalize this approach by learning an embedding for any point on an \textit{object} in the image. 
This still falls short of representing arbitrary points in scenes, and must be retrained in order to adapt to new class of objects.

As a result, keypoint-based approaches can perform very well in specific settings, but act as an extra barrier to deploy robotic systems on novel tasks.
In this work, we aim to demonstrate that recent advances on point tracking models such as~\cite{doersch2023tapir} are good enough to act as a sole perception model for both motion tracking, cross scene correspondence and segmentation.
This system doesn't use any depth information during training or evaluation and uses only a single non-calibrated gripper mounted camera.

Lastly, the ``what'' / ``where'' factorization we explore in this paper has several precedents in the robotics literature, \eg \cite{shridhar2022cliport}, and has been more broadly explored in neuroscience \cite{goodale1998visuomotor,milner2006visual}. 

\subsection{Point Tracking in Computer Vision}
Within computer vision there have been approaches which extract points from any scene without any pretraining.
For example non-learned descriptors such as SIFT~\cite{lowe1999object} or ORB~\cite{rublee2011orb} require no class specific training, but they produce many spurious matches which cannot be easily filtered on non-rigid scenes, making them difficult to use as a target for visual servoing.
Recently there have been several advances in high-performance long-horizon visual tracking such as OmniMotion~\cite{wang2023tracking}, PIPs~\cite{harley2022particle}, TAP-Net~\cite{doersch2022tap} or TAPIR~\cite{doersch2023tapir}, which provide high-quality tracks with very few outliers.
This means that Euclidean distance in point space, with no outlier removal, can serve as a reliable metric of whether two spatial configurations are similar, or whether two motions are similar.
While all of these approaches are made to run on whole videos, we note that the setup (especially TAPIR) can be modified to run online on a frame-by-frame basis, making it suitable for robotics applications.

\section{Approach}
\label{sec:approach}
Recall that we seek to define a \textit{single system} that can be instructed to solve as many manipulation tasks as possible without requiring any per-task engineering or technical knowledge. 
To enable this instruction mechanism we begin by considering the class of semi-parametric policies $\pi(s_t, D)$, in which the policy has access to a ``context dataset'' $D$ at test-time, \eg demonstration trajectories.
Conceptually, RoboTAP can be viewed as a factorization of $\pi$ in which we extract a sufficient statistic for the current action from the full ``context'' $D$, according to the current state $s_t$, \ie $\pi(s_t, D) = \pi(TAP(s_t, q_t), g_t) = \pi(p_t, o_t, g_t)$, where $q_t \in \mathbb{R}^{n, c}$ represents the $c$-dimensional descriptors for $n$ points (a.k.a. the query), $g_t \in Z^{n, 2}$ represents the target locations for those points in the image.
Given a query $q_t$ and an image in the current state $s_t$, \textit{TAP} represents a dense point-tracker that outputs detections $p_t \in Z^{n, 2}$ and an occlusion probability $o_t$ for each point.

The quantities $q_t$ and $g_t$ represent the ``what'' and ``where'', respectively, of the current action, and are extracted from $D$ via a procedure $g_t, q_t=f(s_t, D)$, described in \sref{sec:decomosition}.
For simplicity, in this paper we don't implement $f$ as a per-step operation, but rather as a one-time plan that summarizes the \textit{objectives} common across $D$.
An overview of this procedure and it's connection to the control is illustrated in \fref{fig:overview}.
The high-level outline for our overall solution is as follows:
\begin{enumerate}
    \item Sample large number of query points from $D$ and track those points across all trajectories using TAPIR.
    \item Segment the trajectories into phases, and discover a descriptor-set $q_t$ for each phase which characterize the motion.
    \item Pack the sequence of $q_t$ and corresponding trajectory slices (representing the goals) into a ``motion-plan''.
    \item Execute this motion plan using the low-level controller in a sequential fashion, advancing stages based on a simple final-error criterion.
\end{enumerate}

\subsection{Temporal and Spatial Decomposition}
\label{sec:decomosition}
The first core step of the RoboTAP approach, is to extract important motion from demonstrations in order to construct a motion-plan.
This seen in \fref{fig:overview}.
In order to accomplish this
we assume that our demonstrations $D$ are comprised of an unknown but equal number of phases, each involving the motion of a particular set of points.
Although the set of points could be learned end-to-end (e.g. by behavioral cloning), this would require more data than we have available.  
Therefore, we instead explicitly identify motion segments that are shared across demos, and then discover a set of points for each that are moving consistently, which we term \textit{active points}.

Temporal alignment has been well-studied~\cite{dwibedi2019temporal, davchev2021wish, zakka2022xirl}, but for simplicity, we rely on the assumption that our tasks consist of grasping and releasing objects or making contacts, which means that temporal segments are trivially obtained by thresholding gripper actions and forces.
 
\label{sec:essential_object_selection}
\begin{figure}[t]
  \centering
    \includegraphics[width=0.5\textwidth]{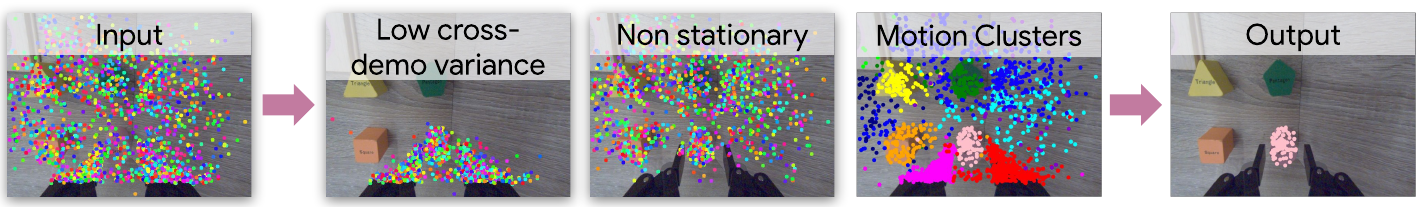}
    \caption{
       \small Active point selection. We exploit the funneling-nature of control to identify relevant points based on their variance, and remove the gripper by filtering static points. Remaining points are used to vote on the \textit{motion cluster}, which we use to sample 128 points throughout the motion-segment to serve as the salient features for that step of the motion plan.
}
  \label{fig:salient_point}
  \vspace{-0.5cm}
\end{figure}

\textbf{Active Point Selection.}
Given a temporal segment, identifying active-points involves asking what points $q_t \in Q$ the low-level controller $\pi_l$ should move in order to generate motions that accomplish a similar result as observed in the segment.
Our criteria for point selection are illustrated in Fig.~\ref{fig:salient_point}.
The key insight is that the active points may begin at diverse locations, but for goal-directed behavior they tend to \textit{end up} the same place across all demos at the end of the relevant segment.
Therefore, we select points which end at the same place at the end of the demo segment (relative to the camera), and remove points which don't move at all across the segment (i.e., the gripper and any already grasped object).  

While this alone can identify many of the desired active points, not all of the useful points are guaranteed to be visible at the end of each motion segment, either due to occlusion or simply failures by the detector; furthermore, there may be inconsistencies due to imprecise demos that make it difficult to set thresholds on whether a point is moving.  
Therefore, we recast the active point discovery as votes on which \textit{object} (or object-part) is being manipulated, which we substantially improves robustness.
However, this means we must perform unsupervised object discovery from the demos, a classically difficult problem which we find is rendered surprisingly reliable given robust dense point tracks.  
We then combine these heuristics with a voting-based scheme, whereby we select the clusters which contains the most points which are ``active'' according to the first two criteria.

\textbf{Clustering.}
There are numerous approaches to object-based clustering, ranging from semantic segmentation~\cite{long2015fully} to generative modeling~\cite{locatello2020object}, but for simplicity, we use motion estimates extracted from TAPIR, since this means we do not require semantic labels, and we find it can work reliably from a remarkably small amount of data.

We first randomly select many points from the image, and track all of them using our online TAPIR model.  We assume that all of the points belong to one of several approximately-rigid objects in the scene, and can therefore be explained by a set of 3D motions followed by reprojection (note that we model camera motion as a motion of all objects in the scene).  Let $p_{i,t}$ be the TAPIR-predicted location for point $i$ at time $t$ in the demos (for simplicity, $t$ indexes both time and demos, i.e., we concatenate all demos into a single long sequence), and $v_{i,t}\in\{0,1\}$ be a thresholded version of occlusion probability $o_t$: 1 if $o_t$ is less than 0.5, 0 otherwise.  We assume there are $K$ different rigid objects in the scene.  The model infers a 3D location $P_{i,k}\in \mathbb{R}^{3}$ for $i$'th point in the $k$'th object (initially, the model does not know which object the point should be assigned to).  It also proposes a rigid 3D transformation for each object at each time $A_{t,k}\in SE(3)$, represented as a $3\times4$ matrix.  We then optimize the transforms and the 3D points to minimize the simple squared error of the reprojection $R(x)=[x[0]/x[2],x[1]/x[2]]$ as follows:

$$\argmin_{A,P}\min_{k}\sum_{i,t}v_{i,t}\|R(A_{t,k}P_{i,k})-p_{i,t}\|^{2}$$

The simplicity of this equation is somewhat remarkable, as it is essentially standard bundle adjustment~\cite{triggs2000bundle}, but \textit{we do not model outliers}.  Outlier rejection is a critical component of prior structure-from-motion methods as they are typically based on descriptors like SIFT~\cite{lowe2004distinctive}, which have a high proportion of extreme errors.  This severely limits their ability to do multi-object reconstruction, as it is difficult to distinguish between small objects and outliers.  TAPIR, however, has powerful occlusion estimation, meaning that all predicted points should be modeled.  We informally tried approaches which used robust losses such as $L1$, Huber, and even truncated $L2$ losses with appropriate RANSAC-style proposals, but did not find formulations that performed on par with simple $L2$.

We parameterize both $P_{i,k}$ and $A_{t,k}$ using neural networks, which aim to capture the inductive biases that points nearby in 2D space, and also frames nearby in time, should have similar 3D configurations.  Specifically, $P_{i,k}=\boldsymbol{P}(p_{i},o_{i}|\theta_1)_{k}$, where $\theta_1$ parameterizes the neural network $\boldsymbol{P}$ which outputs a $k\times 3$ matrix, and $A_{t,k}=\boldsymbol{A}(\phi_{t}|\theta_2)_{k}$, where $\phi_{t}$ is a temporally-smooth learned descriptor for frame $t$, $\theta_2$ is a neural network parameter for neural network $\boldsymbol{A}$ which outputs a $k\times 3 \times 4$ tensor representing rigid transforms.  

The above optimization is, unsurprisingly, somewhat difficult due to local minima; we find that these local minima can be avoided by \textit{splitting} clusters during optimization.  For details on this and the neural networks, see \ifarxiv Appendix \sref{sec:appendix_clustering}. \else our project wepbpage.\fi
Given a clustering, we select clusters by voting.
Every previously selected active point casts a vote for a cluster and we merge clusters with largest number of votes.
To further avoid selecting points from the gripper we compute average point movement for each cluster and remove clusters where this movement is below a threshold.
Finally, we discard any points are not visible in the current phase, or which are not nearby other points on most frames, as these are likely to be tracking failures.
For details, see \ifarxiv Appendix \sref{sec:appendix_hypers}. \else our project wepbpage.\fi

\subsection{Robot controller}
\label{sec:robot_controller}
The foundation of RoboTAP is a general-purpose controller that can align points in a scene.
To define this controller we consider a set of image points, with current locations $p_t$, and goal locations $g_t$ as derived above.
The objective of this controller is to move the gripper with velocity $v_t$ (4 DoF: $\dot{x},\dot{y},\dot{z},\dot{rz}$) such that the points will move towards the desired location, while also being robust to noise and occlusions.
To do this we need an estimate of how the points will move if the gripper moves in a certain direction.
For a wrist-camera this is simply the image Jacobian $\frac{dp}{d\xi} \in \mathbb{R}^{n, 2, 4}$, where $n$ denotes the number of points and $\xi$ contains the relevant dimensions ($x$, $y$, $z$, $rz$) of the gripper pose.

In theory, to obtain the correct image Jacobian we need to consider the camera intrinsics and the extrinsics.
However in our case we only require that it points in the correct direction, and we then tune the rotation and translation gains of the controller for stability.
To demonstrate this in all of our experiments we assume vertical field of view of $90^{\circ}$ and unit depth which leads the following image Jacobian for a single point:
\begin{equation}
J = 
\begin{bmatrix}
1 & 0 & -u & -v \\ 
0 & 1 & -v & u 
\end{bmatrix}
\label{eq:jacobian}
\end{equation}
Where the columns correspond to end-effector-frame linear and angular velocity of the gripper, and u, v are normalized image coordinates such that vertical dimension is within [-1, 1].
Using this Jacobian, we then compute the gripper motion that would by minimize the L2 error between the current detections $p$ and goal locations $g$ under the linear approximation, following standard visual servoing.  
Because the error is typically dominated by translation, applying the Jacobian naively would result in the controller explaining translation via rotation and scaling; therefore, we use Gram-Schmidt orthogonalization~\cite{bjorck1967solving} to eliminate the average translation before computing the Jacobian with respect to rotation and scaling.

If TAPIR provided perfect point locations, the above algorithm would work effectively without modification.
However, for precise tasks, TAPIR errors can still lead to two specific failure modes.
First, outliers from TAPIR can overwhelm the controller when true errors are low.
To deal with this, we leverage TAPIR's uncertainty outputs, and only use points which are confidently predicted for both the current frame and in the demo.
Second, noisy detections introduce a statistical bias towards minimizing the spread of the point cloud in order to reduce unexplainable errors.
This presents itself as a bias towards moving the camera further back.
Therefore, we modify the controller by leveraging the relation between the problems of aligning $p$ to $g$ and the inverse alignment of $g$ to $p$.
This changes the $z$-axis update to perform variance matching rather than directly optimizing the visual servoing objective.
To further reduce variance near the end of each servoing stage (when the demo locations are assumed to agree), we use an average of point locations across all demos as a target to further reduce variance, while we use a single demo as a target earlier in each stage.
\ifarxiv Further details can be found in Appendix \sref{sec:controller_details}. \else See our project webpage for details. \fi

\subsection{Online TAPIR}
\label{sub:online_tapir}
In the previous section we describe how to use TAPIR point tracks inside a robot controller.
However one of the hurdles we must overcome to do this, is to find a way to run it online, inside a control-loop. 
The original TAP-Vid was proposed as an offline benchmark, in that methods may process the entire video before computing a trajectory, and top-perfoming methods like TAPIR~\cite{doersch2023tapir}, PIPs~\cite{harley2022particle}, OmniMotion~\cite{wang2023tracking}, and TAP-Net~\cite{doersch2022tap} all process a full video at once.  
Therefore, a key contribution is to reformulate the top-performing models (i.e. TAPIR) to work online without harming its accuracy.

We first note that TAPIR can be divided into three stages: 1) extracting \textit{query features} for query points, 2) initializing the track via global search on every frame, and 3) refinement with a temporal ConvNet.  To address the first stage, we note that the query features depend only on the query frame, so it is straightforward to extract query feature computation into a separate function.  For the second stage, the initialization for a given frame depends only on the query features and the features for that frame, meaning the initialization can be computed one frame at a time.  

The third stage, however, is a refinement, which is not straightforward to run online: at its core is a convolutional neural network which runs across \textit{time}.  The solution is to observe that this can be converted into an online model by replacing the temporal convolutions with similarly-shaped \textit{causal} convolutions~\cite{oord2016wavenet}, such that the activation of each unit at time $t$ depends \textit{only} on the activations from times $\leq t$.  At training time, this is a trivial architectural change.  When running on the robot, however, it means we must keep a history of input activations for each layer with a temporal receptive field.  
Any activations that a unit at time $t$ depends on are read from the history.  Once the forward pass is complete, a new history is constructed and returned which can be used at the next timestep.  In practice, we find that the performance impact of moving to a causal model is minimal (see tables \ref{tab:query_first_compare} and \ref{tab:robotap_compare}).  See \ifarxiv Appendix~\ref{appendix_causal} \else our project webpage \fi for details.
\section{EVALUATION}

In this section, we aim to demonstrate the strengths and limitations of our system and its components.
The capabilities of RoboTAP overall are directly linked to the precision of TAPIR itself on the domain we are interested in.
Therefore to evaluate and advance its abilities we introduce a new point tracking dataset focused on robotics settings.
Further, since the TAPIR model we are using has been modified to be run in an online way, we evaluate the effect of these changes on it's performance on the TAP-Vid benchmarks.

Following this, we show evaluations of the full RoboTAP system.
In particular, we show that RoboTAP can tackle complex tasks with many stages from just a handful of demos, can deal with start and end poses significantly different from those in the demos, can both place objects precisely and follow trajectories precisely,  has strong invariance to distractors in the workspace, and can deal with non-rigid objects.

We choose a variety of tasks involving placement, insertion, and gluing, that are representative of real-world assembly problems.
While all of the demonstrations were gathered without occlusions and distractor objects we explore how the ability of RoboTAP is affected when these are present.
Finally we also study the servoing precision using a calibrated real-world setup.
\ifarxiv Further experimental ablations of the controller in simulation can be found in Appendix \sref{sec:simulated_ablations}. \fi

\subsection{Robotics TAP-Vid Dataset addition}

The ability to track and relate points in the scene is what enables RoboTAP to generalize to novel scenes and poses.
These capabilities are directly powered by the performance of these models and therefore a core part of our contribution is enabling advances in these areas.
To enable better research we introduce an addition to the previous TAP-Vid benchmark in a form of a new dataset which focuses specifically on robotic manipulation.

Specifically we collect 265 real world robotics manipulation videos. The data source mainly comes from teleoperated episdoes in the public DeepMind robotics videos~\cite{cabi2019scaling,vecerik2020s3k}. We annotate each video with human groundtruth point trajectories, following the same instruction in the previous TAP-Vid dataset~\cite{doersch2022tap}, with 5 points per object and 5 points on the background. To make the evaluation closer to the realistic manipulation setup, we sampled the videos from different camera viewpoints (i.e. basket view, wrist view) where basket view is static and wrist view moves along with the gripper.

\tref{tab:robotap} shows the overall statistics of the newly collected RoboTAP dataset. Comparing to the existing simulated TAP-Vid-RGB-Stacking dataset, it contains more videos, more points, and more frames on average, but more promisingly real world. \tref{tab:points} shows more details of the annotated points in different camera setups. Besides the general evaluation, we further provide labels of whether camera is static or moving and a point is static or moving.

\begin{table}[t]
\resizebox{\linewidth}{!}{
\begin{tabular}{lcc}
\toprule
Dataset & TAP-Vid-RGB-Stacking & RoboTAP \\
\midrule
$\#$ Videos & 50 & 265 \\
$\#$ Points & 30 & 43.7 \\
$\#$ Frames & 250 & 271.9 \\
$\#$ Human Clicks & N/A & 8.5 \\
Sim/Real & Sim & Real \\
Eval resolution & 256x256 & 256x256 \\
\bottomrule
\end{tabular}
}
\caption{
\small Statistics of RoboTAP dataset. Comparing to existing point tracking dataset on Robotics: TAP-Vid-RGB-Stacking, RoboTAP dataset produces realistic challenging videos with more point annotations and longer video durations. Note, $\#$ Points is average number of annotated points per video; $\#$ Frames is the average number of frames per video. $\#$ Human Clicks is the average number of human clicks per video.
}
\label{tab:robotap}
\end{table}

\begin{table}[t]
    \centering
    \begin{tabular}{c c c r}
    \toprule
     & \textbf{Static camera}  & \textbf{Moving camera}  &\textbf{Total} \\
    \midrule
    \textbf{Num total videos} & 103 & 162 & 265 \\
    \textbf{Num total points} & 4441 & 7151 & 11592 \\
    \textbf{Num static points} & 2883 & 1068 & 3951 \\
    \textbf{Num moving points} & 1558 & 6083 & 7641 \\
    \bottomrule
    \end{tabular}
    \caption{\small Point tracks available in the RoboTAP dataset, split by point and camera motion. Note that there are static point tracks in the moving camera videos due to the gripper moves with camera together.}
    \label{tab:points}
    \vspace{-0.5cm}
\end{table}

\subsection{Online TAPIR}

We compare our online TAPIR model with existing baselines on both the TAP-Vid benchmark and the new RoboTAP dataset. Results in \tref{tab:query_first_compare} show that our online TAPIR achieves accuracy similar to the state-of-the-art offline TAPIR model, significantly outperforming TAP-Net and PIPs. On RoboTAP, it achieves an average jaccard (AJ) of 59.1, nearly matching offline TAPIR's 59.6. \tref{tab:robotap_compare} shows more detailed evaluation on different dataset splits.

\begin{table}[t]
\resizebox{\linewidth}{!}{%
\begin{tabular}{l|cccc}
\toprule
 & TAP-Vid-Kinetics & TAP-Vid-DAVIS & TAP-Vid-RGB-Stacking & RoboTAP \\
\midrule
TAP-Net & 39.5 & 36.0 & 50.1 & 45.1 \\
PIPs & 26.1 & 42.5 & - & 29.5 \\
TAPIR & \textbf{52.4} & \textbf{57.0} & 64.9 & \textbf{59.6} \\ \midrule
Online TAPIR & 51.5 & 56.7 & \textbf{67.7} & 59.1 \\ 
\bottomrule
\end{tabular}%
}
\caption{\small Comparison of Online TAPIR to prior results on TAP-Vid and RoboTAP datasets under query first Average Jaccard (AJ) metrics, which considers both position and occlusion accuracy. All models are evaluated under 256x256 resolution. Higher is better.}
\label{tab:query_first_compare}
\end{table}
\begin{table}[t]
\resizebox{\linewidth}{!}{%
\begin{tabular}{l|cccc}
\toprule
& Static Camera & Moving Camera & Static Points & Moving Points \\
\midrule
TAP-Net & 57.0 & 37.8 & 61.3 & 34.5 \\
PIPs & 35.4 & 25.9 & 40.4 & 23.8 \\
TAPIR & \textbf{72.1} & \textbf{52.0} & \textbf{78.3} & \textbf{49.0} \\ \midrule
Online TAPIR & 71.8 & 51.4 & 77.7 & 48.5 \\
\bottomrule
\end{tabular}%
}
\caption{\small Comparison of Online TAPIR to prior results on RoboTAP datasets with more detailed splits. Higher is better.}
\label{tab:robotap_compare}
\vspace{-0.5cm}
\end{table}

\subsection{Robot Setup}
To run our system in the real world we use the Franka Panda Emika robot in the impedance control mode, 2f-85 Robotiq gripper and FT 300 Force Torque Sensor. 
The control signal was sent at 10Hz and interpreted as velocity of the impedance controller setpoint.
To gather images we used a Basler Dart daA1280-54ucm at a resolution of 640x480 pixels mounted to the wrist using a custom 3D printed adapter.
Our camera images were undistorted, but we did not use intrinsic or extrinsic calibration of our camera.

To collect the demonstrations we move the robot via the Franka Panda cuff.
RoboTAP is compatible with kinesthetic teaching because, unlike much other imitation learning work, it does not require access to actions beyond gripper opening/closing state.
During teaching we record robot positions, forces from a wrist force-torque sensor, and wrist camera videos at 10Hz.
We used a keyboard to open and close the gripper.
It takes about 30 seconds to record a demonstration for a pick and place task, 1 minute for the gluing task and up to 2 minutes for the 4-object insertion task.

\begin{figure}[t]
  \centering
  \includegraphics[width=0.49\textwidth]{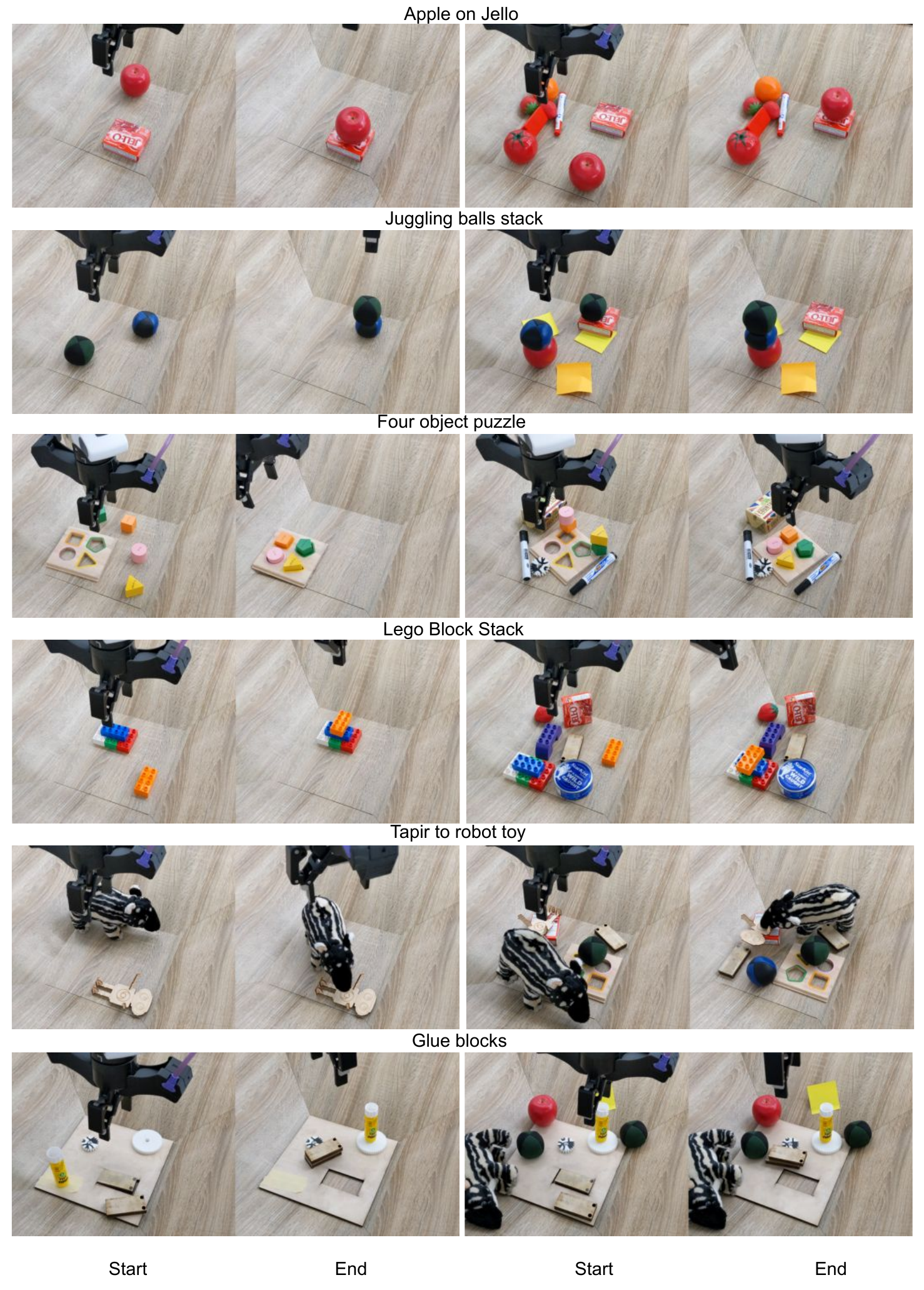}
  \caption{
\small Examples of successfully solved real robot tasks.
In order to challenge the system and demonstrate its robustness we show its performance on scenes with clutter, distracting objects and partial occlusions.
}
  \label{fig:real_examples}
  \vspace{-0.2cm}
\end{figure}

\begin{table}[t]
\Huge
\begin{center}
\resizebox{0.49\textwidth}{!}{
\begin{tabular}{c|c|c}
\toprule
\textbf{Task name} & \textbf{Description} & \textbf{$\#$ Demos} \\
 \midrule
 \Huge Apple on jello & Pick up the apple and place it on jello. & 5 \\
 \midrule
 \Huge Juggling ball stack & Pick up a green juggling ball and place it on a blue ball. & 3 \\
 \midrule
\Huge Four object puzzle & Place four wooden objects into their cutouts on a puzzle board. & 6 \\
 \midrule
 \Huge Lego Stack & Put orange LEGO brick on top of a blue brick and push them together. & 4 \\
 \midrule
\Huge Tapir to robot & Pick up the plush tapir toy and place it next to the wooden robot. & 4\\
 \midrule
\Huge Gluing & Glue the 2 wooden block together and place them beside the white gear. & 5\\
 \midrule

  \Huge Gear on grid & Pick up the white gear and place it on the grid. & 6 \\
 \midrule
 \Huge Four object stack & Stack 4 wooden objects on top of each other. & 4\\
 \midrule
 \Huge Pass the butter & Pick up a butter and place it in human's hand. & 5 \\
 \midrule
\bottomrule
\end{tabular}}
\end{center}
\vspace{-0.3cm}
\caption{\small Tasks details.}
\label{tabl:tasks_descriptions}
\vspace{-0.5cm}
\end{table}
We used a total of 9 tasks described in Table~\ref{tabl:tasks_descriptions}, some of which are illustrated in Figure~\ref{fig:real_examples} - see our \href{https://robotap.github.io}{website} for full list of task videos.
When running this controller we have noticed a very repeatable pattern of cases where the controller succeeds and where it fails.
For a majority of the tasks we have attempted we observe robust performance provided the algorithm's basic assumptions are met,
yet we have seen it fail in a few specific cases.
Therefore we believe that a simple success metric would not appropriately capture its behaviour and we aim to provide other avenues to express its performance.

In all cases we aimed to demonstrate the performance of the controller in a clean environment, and then show how performance degrades with increased clutter and occlusions.
The only tasks where we have not observed reliable performance were the LEGO stack, in which the controller lacked the precision to stack the bricks, and the 4 object stack, where the compounding of the errors often lead the last object to be dropped in the wrong location.
Overall we observed 4 factors which caused failures:
\emph{(1) Occlusions} cause failures when the active points cannot be seen.
This most notably happens due to the gripper occlusions.
\emph{(2) Scale changes} cause failures when the gripper moves too far away and none of the currently-tracked features can be matched because the objects are too small.
\emph{(3) Distractors} can cause failures when the scene contains a similar object to the essential object such as a red apple vs a tomato.
\emph{(4) Collisions} can happen with the gripper or currently-held object as the controller does not have a way to perceive clutter.
However in absence of these cases, we observed robustness to novel arrangements and even dynamic changes of the scene while the controller is being executed.

\begin{figure}[t]
  \centering
  \vspace{0.2cm}
  \includegraphics[width=0.5 \textwidth]{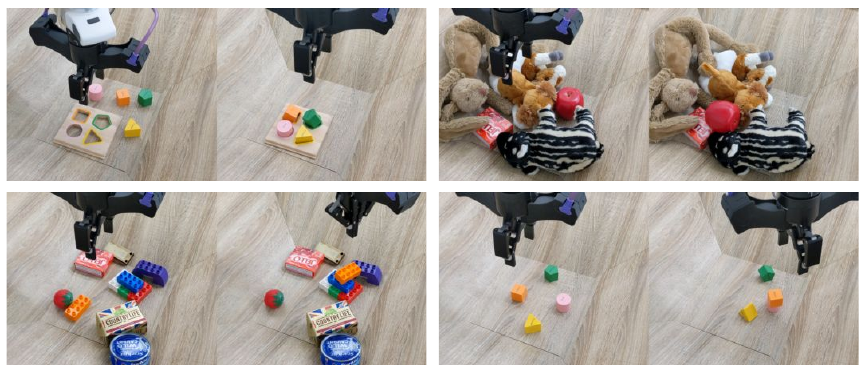}
  \caption{
  \small Examples of states where the system failed to reach the desired final state.
  Tasks which require sub 5mm precision cannot always be reliably solved (\eg shape-matching).
  In addition, our use of a purely-visual control paradigm makes it difficult to solve tasks that require reasoning over visual and force modalities simultaneously (\eg lego part-mating).
  Lastly, our controller is unable to reason about the validity of a motion-plan at runtime, which can lead to failures if certain motions are invalid (\eg the bunny's ear is covering the jello too much to place the apple).
}
  \label{fig:failure_examples}
  \vspace{-0.5cm}
\end{figure}

\subsection{Real world precision}
\label{sec:precision_placement}
Previous experiments demonstrated the qualitative behavior of several long-horizon manipulation tasks.
We conclude by evaluating the precision of our visual-servoing controller quantitatively.
For this we take inspiration from position-repeatability analyses performed on commercial industrial robot arms.
We gathered 6 demonstrations of putting a textured white plastic gear on a square of graph paper, which allowed us to precisely compute of the placement error using OpenCV \cite{bradski2000opencv}.
For the demonstrations we used 3 different configurations of the white gear and the target pattern.
The target had a cross in the middle which allowed us to take pictures of the centre of the gear and automatically extract its location.

Then we ran the controller 30 times for each of 3 novel goal locations.
First goal location, \textit{Near}, was positioned in the middle between the locations which were used for demonstrations.
The second one, \textit{Far}, was on the opposite side of our workspace and the last one, \textit{Rotated}, was on the side but rotated by 90 degrees.
Notably the second 2 goal positions are outside of the distribution provided during demonstrations.

\begin{figure}[t]
  \centering
  \vspace{0.2cm}
    \includegraphics[width=0.35\textwidth]{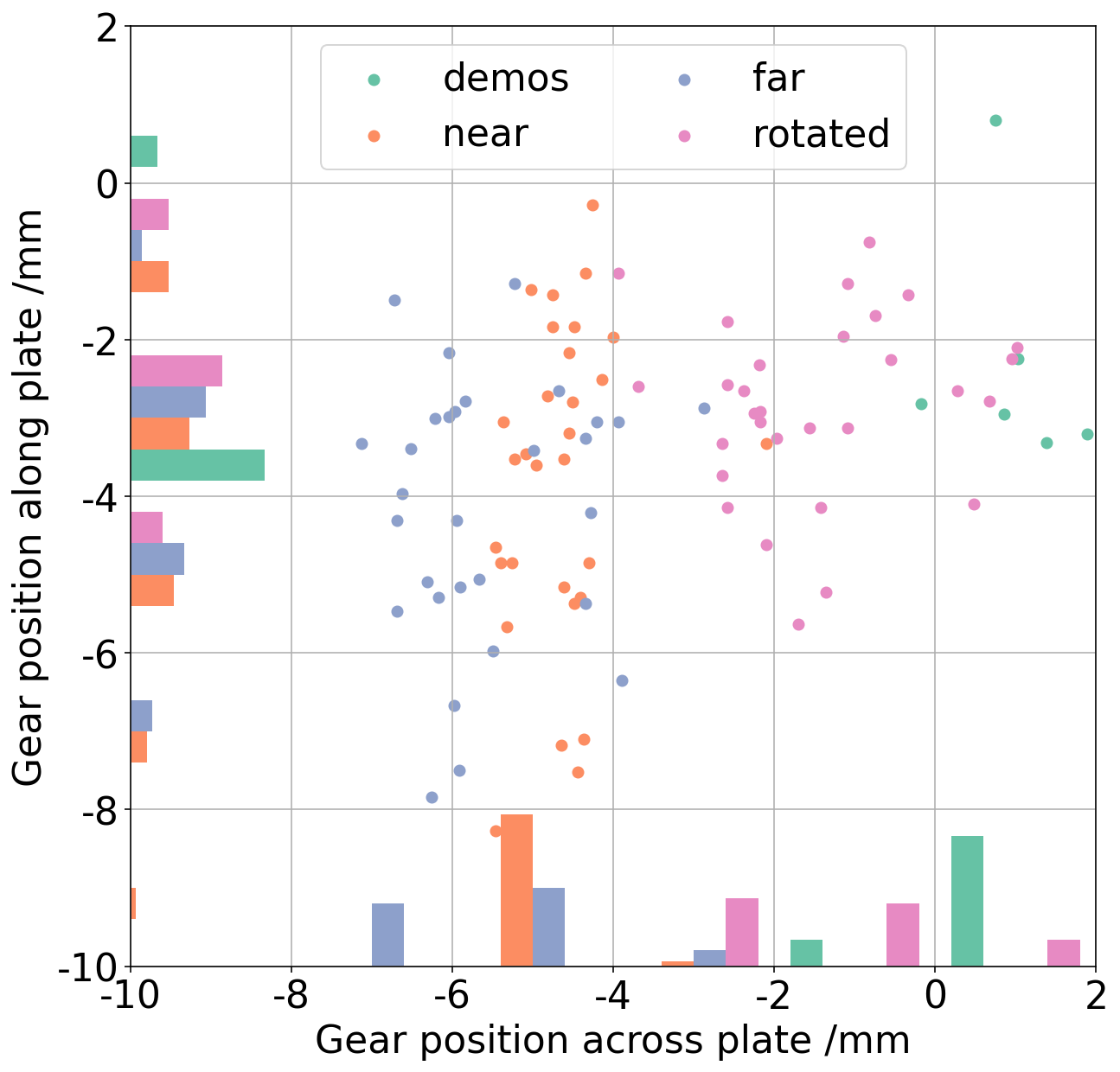}
    \caption{
\small Evaluation of the precision of controller.
We designed a pick-and-place task which allowed us to easily measure exact final location of the placed object relative to movable target.
We gathered 6 demonstrations and ran RoboTAP 30 times for 3 different locations of the placement target.
The figure shows the locations of where the object was placed in each of the trials.
When the goal was located between the demonstrated position the distribution is similar to what was observed during the demonstrations ($<4$mm).
We can see a slight decrease in precision when the target was moved to a novel location or rotated by 90 degrees ($<1$cm).
}
  \label{fig:precision_placement}
  \vspace{-0.5cm}
\end{figure}

We can see the results of this evaluation in Fig.~\ref{fig:precision_placement}.
Based on the figure we can see that the error of the controller is comparable to the error seen across the demonstrations.
The error along the plate, orthogonal to the grasp direction is on the order of several milimeters.
Based on these results we can see that the controller in its current form would not be suitable for sub-mm insertion tasks, however we have not attempted to extensively tune it for this purpose.
Further information about these experiments is provided in Appendix \sref{sec:appendix_precision_placement}.

\section{CONCLUSION}
We presented RoboTAP, a manipulation system that can solve novel visuomotor tasks from just a few minutes of robot interaction.
RoboTAP does not require any task-specific training or neural-network fine-tuning. 
Thanks largely to the generality of TAP, we found that adding new tasks (including tuning hyper-parameters) took minutes, which is orders of magnitude faster than any manipulation system we are familiar with.
We believe that this capability could be useful for large-scale autonomous data-gathering, and perhaps as a solution real-world tasks in its own right.
RoboTAP is most useful in scenarios where quick teaching of visuomotor skills is required, and where it is easy to demonstrate the desired behavior a few times.

There are several important limitations of RoboTAP.
First, the low-level controller is purely visual, which precludes complex motion planning or  force-control behavior. 
Second, we currently compute the motion plan once and execute it without re-planning, which could fail if individual behaviors fail or if the environment changes unexpectedly.

Several of the ideas in RoboTAP, \eg explicit spatial representation and short-horizon visuo-motor control, could also be applicable in more general settings.
In the future we would like to explore whether RoboTAP models and insights can be combined with larger-scale end-to-end models to increase their efficiency and interpretability.

\section*{ACKNOWLEDGMENT}
We would like to thank Tom Roth{\"o}rl, Akhil Raju, and Marlon Gwira for their help with our robot setup and Dilara Gokay for dataset infrastructure support.
We would also like to thank Nicolas Heess, Oleg Sushkov, Andrew Zisserman and João Carreira for useful discussions and feedback.

\bibliographystyle{IEEEtran}
\bibliography{IEEEfull.bib}

\begin{thebibliography}{10}
\providecommand{\url}[1]{#1}
\csname url@rmstyle\endcsname
\providecommand{\newblock}{\relax}
\providecommand{\bibinfo}[2]{#2}
\providecommand\BIBentrySTDinterwordspacing{\spaceskip=0pt\relax}
\providecommand\BIBentryALTinterwordstretchfactor{4}
\providecommand\BIBentryALTinterwordspacing{\spaceskip=\fontdimen2\font plus
\BIBentryALTinterwordstretchfactor\fontdimen3\font minus
  \fontdimen4\font\relax}
\providecommand\BIBforeignlanguage[2]{{%
\expandafter\ifx\csname l@#1\endcsname\relax
\typeout{** WARNING: IEEEtran.bst: No hyphenation pattern has been}%
\typeout{** loaded for the language `#1'. Using the pattern for}%
\typeout{** the default language instead.}%
\else
\language=\csname l@#1\endcsname
\fi
#2}}

\bibitem{doersch2022tap}
C.~Doersch, A.~Gupta, L.~Markeeva, A.~Recasens, L.~Smaira, Y.~Aytar,
  J.~Carreira, A.~Zisserman, and Y.~Yang, ``Tap-vid: A benchmark for tracking
  any point in a video,'' \emph{Proceedings of Neural Information Processing
  Systems (NeurIPS)}, 2022.

\bibitem{saycan2022arxiv}
A.~Michael~et al, ``Do as i can and not as i say: Grounding language in robotic
  affordances,'' in \emph{arXiv:2204.01691}, 2022.

\bibitem{bousmalis2023robocat}
K.~Bousmalis, G.~Vezzani, R.~Dushyant, C.~Devin, A.~X. Lee, M.~Bauza,
  T.~Davchev, Y.~Zhou, and A.~Gupta~et al, ``Robocat: A self-improving
  foundation agent for robotic manipulation,'' \emph{arXiv:2306.11706}, 2023.

\bibitem{brohan2022rt}
A.~Michael~et al, ``{RT}-1: Robotics transformer for real-world control at
  scale,'' \emph{arXiv:2212.06817}, 2022.

\bibitem{liang2023code}
J.~Liang, W.~Huang, F.~Xia, P.~Xu, K.~Hausman, B.~Ichter, P.~Florence, and
  A.~Zeng, ``Code as policies: Language model programs for embodied control,''
  in \emph{ICRA}.\hskip 1em plus 0.5em minus 0.4em\relax IEEE, 2023, pp.
  9493--9500.

\bibitem{kalashnikov2018qt}
D.~Kalashnikov, A.~Irpan, P.~Pastor, J.~Ibarz, A.~Herzog, E.~Jang, D.~Quillen,
  E.~Holly, M.~Kalakrishnan, V.~Vanhoucke, \emph{et~al.}, ``Qt-opt: Scalable
  deep reinforcement learning for vision-based robotic manipulation,''
  \emph{arXiv:1806.10293}, 2018.

\bibitem{yen2020learning}
L.~Yen-Chen, A.~Zeng, S.~Song, P.~Isola, and T.-Y. Lin, ``Learning to see
  before learning to act: Visual pre-training for manipulation,'' in
  \emph{ICRA}.\hskip 1em plus 0.5em minus 0.4em\relax IEEE, 2020, pp.
  7286--7293.

\bibitem{lee2020making}
M.~A. Lee, Y.~Zhu, P.~Zachares, M.~Tan, K.~Srinivasan, S.~Savarese, L.~Fei-Fei,
  A.~Garg, and J.~Bohg, ``Making sense of vision and touch: Learning multimodal
  representations for contact-rich tasks,'' \emph{T:RO}, vol.~36, no.~3, pp.
  582--596, 2020.

\bibitem{luo2021robust}
J.~Luo, O.~Sushkov, R.~Pevceviciute, W.~Lian, C.~Su, M.~Vecerik, N.~Ye,
  S.~Schaal, and J.~Scholz, ``Robust multi-modal policies for industrial
  assembly via reinforcement learning and demonstrations: A large-scale
  study,'' \emph{arXiv:2103.11512}, 2021.

\bibitem{zhu2014single}
M.~Zhu, K.~G. Derpanis, Y.~Yang, S.~Brahmbhatt, M.~Zhang, C.~Phillips,
  M.~Lecce, and K.~Daniilidis, ``Single image 3d object detection and pose
  estimation for grasping,'' in \emph{ICRA}.\hskip 1em plus 0.5em minus
  0.4em\relax IEEE, 2014, pp. 3936--3943.

\bibitem{xiang2017posecnn}
Y.~Xiang, T.~Schmidt, V.~Narayanan, and D.~Fox, ``Posecnn: A convolutional
  neural network for 6d object pose estimation in cluttered scenes,''
  \emph{arXiv:1711.00199}, 2017.

\bibitem{deng2020self}
X.~Deng, Y.~Xiang, A.~Mousavian, C.~Eppner, T.~Bretl, and D.~Fox,
  ``Self-supervised 6d object pose estimation for robot manipulation,'' in
  \emph{ICRA}.\hskip 1em plus 0.5em minus 0.4em\relax IEEE, 2020, pp.
  3665--3671.

\bibitem{chen2022sim}
K.~Chen, R.~Cao, S.~James, Y.~Li, Y.-H. Liu, P.~Abbeel, and Q.~Dou,
  ``Sim-to-real 6d object pose estimation via iterative self-training for
  robotic bin picking,'' in \emph{ECCV}.\hskip 1em plus 0.5em minus 0.4em\relax
  Springer, 2022, pp. 533--550.

\bibitem{chen2023texpose}
H.~Chen, F.~Manhardt, N.~Navab, and B.~Busam, ``Texpose: Neural texture
  learning for self-supervised 6d object pose estimation,'' in
  \emph{CVPR}.\hskip 1em plus 0.5em minus 0.4em\relax IEEE/CVF, 2023, pp.
  4841--4852.

\bibitem{hill1979real}
J.~Hill, ``Real time control of a robot with a mobile camera,'' in \emph{Proc.
  9th Int. Symp. on Industrial Robots}, 1979, pp. 233--245.

\bibitem{pomares2019visual}
J.~Pomares, ``Visual servoing in robotics,'' p. 1298, 2019.

\bibitem{vahrenkamp2008visual}
N.~Vahrenkamp, S.~Wieland, P.~Azad, D.~Gonzalez, T.~Asfour, and R.~Dillmann,
  ``Visual servoing for humanoid grasping and manipulation tasks,'' in
  \emph{ICHR}.\hskip 1em plus 0.5em minus 0.4em\relax IEEE-RAS, 2008, pp.
  406--412.

\bibitem{kragic2002survey}
D.~Kragic, H.~I. Christensen, \emph{et~al.}, ``Survey on visual servoing for
  manipulation,'' \emph{Computational Vision and Active Perception Laboratory,
  Fiskartorpsv}, vol.~15, p. 2002, 2002.

\bibitem{ribeiro2021real}
E.~G. Ribeiro, R.~de~Queiroz~Mendes, and V.~Grassi~Jr, ``Real-time deep
  learning approach to visual servo control and grasp detection for autonomous
  robotic manipulation,'' \emph{RAS}, vol. 139, p. 103757, 2021.

\bibitem{manuelli2019kpam}
L.~Manuelli, W.~Gao, P.~Florence, and R.~Tedrake, ``kpam: Keypoint affordances
  for category-level robotic manipulation,'' in \emph{The International
  Symposium of Robotics Research}.\hskip 1em plus 0.5em minus 0.4em\relax
  Springer, 2019, pp. 132--157.

\bibitem{vecerik2020s3k}
M.~Vecerik, J.-B. Regli, O.~Sushkov, D.~Barker, R.~Pevceviciute,
  T.~Roth{\"o}rl, C.~Schuster, R.~Hadsell, L.~Agapito, and J.~Scholz, ``S3k:
  Self-supervised semantic keypoints for robotic manipulation via multi-view
  consistency,'' \emph{Conference on Robotic Learning (CoRL)}, 2020.

\bibitem{das2021model}
N.~Das, S.~Bechtle, T.~Davchev, D.~Jayaraman, A.~Rai, and F.~Meier,
  ``Model-based inverse reinforcement learning from visual demonstrations,'' in
  \emph{CoRL}.\hskip 1em plus 0.5em minus 0.4em\relax PMLR, 2021, pp.
  1930--1942.

\bibitem{vecerik2022few}
M.~Vecerik, J.~Kay, R.~Hadsell, L.~Agapito, and J.~Scholz, ``Few-shot keypoint
  detection as task adaptation via latent embeddings,'' in \emph{ICRA}.\hskip
  1em plus 0.5em minus 0.4em\relax IEEE, 2022, pp. 1251--1257.

\bibitem{florence2018dense}
P.~R. Florence, L.~Manuelli, and R.~Tedrake, ``Dense object nets: Learning
  dense visual object descriptors by and for robotic manipulation,''
  \emph{arXiv:1806.08756}, 2018.

\bibitem{doersch2023tapir}
C.~Doersch, Y.~Yang, M.~Vecerik, D.~Gokay, A.~Gupta, Y.~Aytar, J.~Carreira, and
  A.~Zisserman, ``Tapir: Tracking any point with per-frame initialization and
  temporal refinement,'' \emph{arXiv:2306.08637}, 2023.

\bibitem{shridhar2022cliport}
M.~Shridhar, L.~Manuelli, and D.~Fox, ``Cliport: What and where pathways for
  robotic manipulation,'' in \emph{CoRL}.\hskip 1em plus 0.5em minus
  0.4em\relax PMLR, 2022.

\bibitem{goodale1998visuomotor}
M.~A. Goodale, ``Visuomotor control: Where does vision end and action begin?''
  \emph{Current Biology}, vol.~8, no.~14, pp. R489--R491, 1998.

\bibitem{milner2006visual}
D.~Milner and M.~Goodale, \emph{The visual brain in action}.\hskip 1em plus
  0.5em minus 0.4em\relax OUP Oxford, 2006, vol.~27.

\bibitem{lowe1999object}
D.~G. Lowe, ``Object recognition from local scale-invariant features,'' in
  \emph{ICCV}, vol.~2.\hskip 1em plus 0.5em minus 0.4em\relax Ieee, 1999, pp.
  1150--1157.

\bibitem{rublee2011orb}
E.~Rublee, V.~Rabaud, K.~Konolige, and G.~Bradski, ``Orb: An efficient
  alternative to sift or surf,'' in \emph{ICCV}.\hskip 1em plus 0.5em minus
  0.4em\relax Ieee, 2011, pp. 2564--2571.

\bibitem{wang2023tracking}
Q.~Wang, Y.-Y. Chang, R.~Cai, Z.~Li, B.~Hariharan, A.~Holynski, and N.~Snavely,
  ``Tracking everything everywhere all at once,'' \emph{arXiv:2306.05422},
  2023.

\bibitem{harley2022particle}
A.~W. Harley, Z.~Fang, and K.~Fragkiadaki, ``Particle video revisited: Tracking
  through occlusions using point trajectories,'' in \emph{ECCV}.\hskip 1em plus
  0.5em minus 0.4em\relax Springer, 2022, pp. 59--75.

\bibitem{dwibedi2019temporal}
D.~Dwibedi, Y.~Aytar, J.~Tompson, P.~Sermanet, and A.~Zisserman, ``Temporal
  cycle-consistency learning,'' in \emph{Proceedings of the IEEE/CVF conference
  on computer vision and pattern recognition}, 2019, pp. 1801--1810.

\bibitem{davchev2021wish}
T.~Davchev, O.~Sushkov, J.-B. Regli, S.~Schaal, Y.~Aytar, M.~Wulfmeier, and
  J.~Scholz, ``Wish you were here: Hindsight goal selection for long-horizon
  dexterous manipulation,'' \emph{ICLR}, 2022.

\bibitem{zakka2022xirl}
K.~Zakka, A.~Zeng, P.~Florence, J.~Tompson, J.~Bohg, and D.~Dwibedi, ``Xirl:
  Cross-embodiment inverse reinforcement learning,'' in \emph{Conference on
  Robot Learning}.\hskip 1em plus 0.5em minus 0.4em\relax PMLR, 2022, pp.
  537--546.

\bibitem{long2015fully}
J.~Long, E.~Shelhamer, and T.~Darrell, ``Fully convolutional networks for
  semantic segmentation,'' in \emph{CVPR}.\hskip 1em plus 0.5em minus
  0.4em\relax IEEE, 2015, pp. 3431--3440.

\bibitem{locatello2020object}
F.~Locatello, D.~Weissenborn, T.~Unterthiner, A.~Mahendran, G.~Heigold,
  J.~Uszkoreit, A.~Dosovitskiy, and T.~Kipf, ``Object-centric learning with
  slot attention,'' \emph{NeurIPS}, pp. 11\,525--11\,538, 2020.

\bibitem{triggs2000bundle}
B.~Triggs, P.~F. McLauchlan, R.~I. Hartley, and A.~W. Fitzgibbon, ``Bundle
  adjustment—a modern synthesis,'' in \emph{Vision Algorithms: Theory and
  Practice: International Workshop on Vision Algorithms Corfu}.\hskip 1em plus
  0.5em minus 0.4em\relax Springer, 2000, pp. 298--372.

\bibitem{lowe2004distinctive}
D.~G. Lowe, ``Distinctive image features from scale-invariant keypoints,''
  \emph{IJCV}, vol.~60, pp. 91--110, 2004.

\bibitem{bjorck1967solving}
{\AA}.~Bj{\"o}rck, ``Solving linear least squares problems by gram-schmidt
  orthogonalization,'' \emph{BIT Numerical Mathematics}, vol.~7, no.~1, pp.
  1--21, 1967.

\bibitem{oord2016wavenet}
A.~Oord, S.~Dieleman, H.~Zen, K.~Simonyan, O.~Vinyals, A.~Graves,
  N.~Kalchbrenner, A.~Senior, and K.~Kavukcuoglu, ``Wavenet: A generative model
  for raw audio,'' \emph{arXiv:1609.03499}, 2016.

\bibitem{cabi2019scaling}
S.~Cabi, S.~G. Colmenarejo, A.~Novikov, K.~Konyushkova, S.~Reed, R.~Jeong,
  K.~Zolna, Y.~Aytar, D.~Budden, M.~Vecerik, \emph{et~al.}, ``Scaling
  data-driven robotics with reward sketching and batch reinforcement
  learning,'' \emph{arXiv preprint arXiv:1909.12200}, 2019.

\bibitem{bradski2000opencv}
G.~Bradski, ``The opencv library.'' \emph{Dr. Dobb's Journal: Software Tools
  for the Professional Programmer}, vol.~25, no.~11, pp. 120--123, 2000.

\bibitem{todorov2012mujoco}
E.~Todorov, T.~Erez, and Y.~Tassa, ``Mujoco: A physics engine for model-based
  control,'' in \emph{Proc. of IROS}, 2012, pp. 5026--5033.

\bibitem{IgnitionFuel-GoogleResearch-Google-Scanned-Objects}
\BIBentryALTinterwordspacing
GoogleResearch. (2020, September) Google scanned objects. Open Robotics.
  [Online]. Available:
  \url{https://fuel.ignitionrobotics.org/1.0/GoogleResearch/fuel/collections/Google\%20Scanned\%20Objects}
\BIBentrySTDinterwordspacing

\bibitem{sun2021research}
Y.~Sun, J.~Falco, M.~A. Roa, and B.~Calli, ``Research challenges and progress
  in robotic grasping and manipulation competitions,'' \emph{IEEE robotics and
  automation letters}, vol.~7, no.~2, pp. 874--881, 2021.

\bibitem{kuznetsova2021efficient}
A.~Kuznetsova, A.~Talati, Y.~Luo, K.~Simmons, and V.~Ferrari, ``Efficient video
  annotation with visual interpolation and frame selection guidance,'' in
  \emph{Proceedings of the IEEE/CVF Winter Conference on Applications of
  Computer Vision}, 2021, pp. 3070--3079.

\end{thebibliography}

\ifarxiv
\appendices

\section{Simulated ablations}
\label{sec:simulated_ablations}
To explore the choices behind our visual servoing controller and quantitatively evaluate them we define a MuJoCo simulated environment \cite{todorov2012mujoco}.
Within this environment we place a camera and an object from the Google Scanned Objects dataset \cite{IgnitionFuel-GoogleResearch-Google-Scanned-Objects}.
We create a dataset of 480 trajectories with 20 different objects and 24 motions per object.
There trajectories contain changes up to 5x in scale and movements across the whole image.
The task is for the controller to follow a demonstrated trajectory using the point tracks provided by TAPIR.
To make the environment more visually complex we add a spatially fixed background to each trajectory which is randomly sampled from the real demonstration data collected for our tasks.
This acts as a distraction for the TAPIR model and increases the observed noise to better match the real trajectories.

Our visual servoing controller needs a set of points $q$ to be tracked.
In simulation we construct this by generating an extra video with a different background to the demonstration and randomly sample points across different timesteps of the demonstration.
To evaluate our controller we sample a yet different background and match the initial relative camera and object pose to the one seen in the demonstration.
Then at each timestep the controller outputs an action which is a desired pose difference to the next state.
We run this for each of the 480 trajectories and evaluate in how many of the videos the controller converges to the final state of the trajectory.
To evaluate visual servoing with 6 degrees of freedom (DOF) we used the jacobian described in \eqref{eq:jacobian_full} instead of \eqref{eq:jacobian}.
We note that these visual servoing tasks are very challenging as they contain large changes of scale and the tracked object presents only a very small part of the scene.
The results of these experiments are presented in \tref{tab:sim_results}.
First we can see that this approach can work when all 6 degrees of freedom are being used for servoing, however it is significantly less stable.
If we do not evaluate the jacobian at both demonstration and current location, we also observe a significant drop in performance due to the statistical bias.
We see similar but smaller performance drop when we do not orthogonalise the jacobian, \ie if we replace $J^\perp$ with $J$.
Both of these issues are especially notable in cases which contain large changes in scale.
\begin{table}[]
    \centering
    \begin{tabular}{c c}
    \hline
     & \textbf{Simulation success rate} \\
    \hline
    \textbf{Full RoboTAP} & 83.8 +- 1.7 \\
    \textbf{6 DOF} & 37.5 +- 2.2 \\
    \textbf{Single directional jacobian} & 61.3 +- 2.2 \\
    \textbf{No orthogonalisation} & 74.6 +- 2.0 \\
    \hline
    \end{tabular}
    \caption{Ablation of controller parameters in simulation.}
    \label{tab:sim_results}
\end{table}
\section{Clustering Implementation Details}
\label{sec:appendix_clustering}
Recall that we aim to find, for each object, optimal 3D canonical locations $P_{i,k}\in \mathbb{R}^{3}$ for each point, and optimal object pose $A_{t,k}\in SE(3)$ for each frame, which minimizes the error of the reprojection $R(x)=[x[0]/x[2],x[1]/x[2]]$ with respect to predicted points $p_{i,t}$:

$$L(A,P)=\min_{k}\sum_{i,t}v_{i,t}\|R(A_{t,k}P_{i,k})-p_{i,t}\|^{2}$$

Where $i$ indexes points, $t$ indexes video frames, and $k$ indexes the clusters.  We parameterize $P$ and $A$ with simple neural networks.  This problem is somewhat underconstrained and difficult to optimize, so we use neural networks to inject some priors.  

\textbf{Object transforms} We enforce temporal smoothness by making $A$ a function of a smooth latent variable $V \in \mathbb{R}^{T\times128}$.  The smoothness of $V$ is enforced by making it a function of a parameter $v  \in \mathbb{R}^{T\times64}$ convolved with a depthwise convolutional filter of shape $128\times64$ which is learned as a part of the optimization.  $V_t$ is then projected to 128 dimensions and passed to a small residual neural network (2 residual blocks with a bottleneck of 64 units) before a final neural network projection to the parameters of $k$ 3D rigid transforms, such that $A(V_t)$ can be reshaped to $k\times 3\times4$.  We constrain $A$ to be an in-plane rotation (i.e., the first three columns of last row are set to $[0, 0, 1]$, make the first two rows orthogonal to this, make the first two rows orthomormal by projection, and multiply by the determinant; thus the neural network output has shape $k * 7$: 4 values parameterizing rotation and 3 parameterizing position).  Note that because neural network parameters are shared across clusters, the network is likely to output similar transformations for all clusters when this doesn't increase the loss, making it easier for points to move between clusters.

\textbf{3D point locations} We enforce that points that are spatially close and have similar motion should have similar 3D locations.  Thus, we make $P_{i,k}$ a function of $p_i$.  Unfortunately, $p_i$ can have undefined values where points are not visible, which means directly applying an MLP can misbehave.  Therefore, we define a bank of centroid values $C\in \mathbb{R}^{384 \times T \times 2}$ where 384 is the is a fixed constant number of centroids; these are initialized by sampling from the data itself.  To encode a given trajectory $p_i$, we compute the distance $d$ between $p_i$ and all centroids for frames where $p_i$ is visible, normalized by the number of visible frames.  We then apply a Gaussian kernel by computing $exp(-d^{2})$, creating an encoding for each point of size 384.  This encoding is then projected to 64 dimensions, before being passed to a small residual neural network (2 residual blocks with a bottleneck of 32) before being projected to $k\times 3$, i.e., a 3D position hypothesis for every cluster.  Note that because the parameters are shared for all clusters, the network is likely to output similar points for all clusters when this doesn't increase the loss, makign it easier for points to move between clusters.

\textbf{Splitting and merging} Directly optimizing the reprojection error tends to result in the algorithm getting stuck in local minima: motion is dominated by camera motion, and so a single cluster which explains camera motion best will tend to get all points assigned to it, leaving no learning signal for the remaining clusters.  While it is tempting to use some kind of soft assignment, we found that such schemes typically resulted in the remaining clusters explaining just a small number of outliers, rather than discovering independently-moving objects all at once. However, one surprisingly effective approach (of which we are not aware of prior work) is to \textit{recursively split} clusters whenever doing so substantially improves the loss.  To accomplish this, note that only the final linear projection layers of our two neural networks depends on the number of clusters $k$: the parameters for these layers can be written as a matrix $w \in \mathbb{R}^{k\times c}$ for some numner of channels $c$.   For each such weight matrix, we create two new weight matrices $w^{\prime} \in \mathbb{R}^{k\times c}$ and $w^{\prime\prime} \in \mathbb{R}^{k\times c}$, where the $k$'th row parameterizes a new clustering where the $k$'th row of $w$ has been split into two different clusters, which we term `forks' of the original weight matrix.
We compute the loss under every possible split, and optimize for the split with the minimum loss.  Mathematically, define $w^{K} \in \mathbb{R}^{(k + 1)\times c}$ to be a new matrix where the $K$-th row of $w$ has been removed, and the $K$-th rows of both $w^{\prime}$ and $w^{\prime\prime}$ have been appended.  We can use $w^{K}$ to compute two new 3D locations and 3D transformations $A^{K}$ and $P^{K}$.  Then we minimize the following loss:

$$\argmin_{\theta} \min_{K} L(A^{K}(\theta),P^{K}(\theta))$$

\noindent Here, $\theta$ parameterizes the neural networks that output $A$ and $P$, and includes $w$ and both of the `fork' variables  $w^{\prime}$ and $w^{\prime\prime}$.  After a few hundred optimization steps, we replace $w$ with $w^{K}$, and create new `forks` of this matrix (initializing the forks with small perturbations of $w^{K}$).  We begin with $k=1$ and repeating recursive forking process until the desired number of objects is reached.  We find that this recursive splitting can result in over-segmentation, but this can be somewhat mitigated by deleting clusters after optimization is finished.  This is an analogous process: we optimize for the minimum loss after clusters are deleted.  Our full implementation of this algorithm is available in our public project repository.

\section{Implementation details of online tapir}
\label{appendix_causal}
In this section, we provide implementation details for our causal version of the TAPIR model.  Our full implementation can be found in our project Github.

Recall that the temporal point refinement of the TAPIR model uses a depthwise convolutional module, where the query point features, the $x$ and $y$ positions, occlusion and uncertainty estimates, and score maps for each frame are all concatenated into a single sequence, and the convolutional model outputs an update for the position and occlusion. 

The depthwise convolutional model replaces every depthwise layer in the original model with a causal depthwise convolution; therefore, the resulting model has the same number of parameters as the original TAPIR model, with all hidden layers having the same shape.  There is no change to the training procedure, as we can concatenate all frames in the training sequences and run the convolutions across the full sequence.

At test time, however, we must preserve a ``causal context'' across frames, which contains the activations computed for the last frame that are required as input for the depthwise convolution layers.  Because the temporal receptive field of the depthwise is 3, we must keep 2 frames of context for every depthwise convolution.
The refinement steps use 4 PIPs iterations, each of which has 12 blocks, and each block has 2 temporal depthwise conv layers with 512 and 2048 units respectively.  Therefore, the temporal context is $2\times(512+2048)\times 12\times 4=245K$ floating-point values for every point.  

\section{Precision experiments - further information}
\label{sec:appendix_precision_placement}

In this section we present further information about the precision experiments presented in \sref{sec:precision_placement}.
In \fref{fig:precision_examples} we show images of initial states from our experiments.
During the 30 evaluations we did not move the target and always manually returned the gear to a similar position within several centimeters.
In \tref{tab:precision_table} we present the means and spreads of points along the x and y directions from our experiments.
These are the same results as in \fref{fig:precision_placement} to present an alternative quantitative analysis.

\begin{figure}[t]
  \centering
  \includegraphics[width=0.49\textwidth]{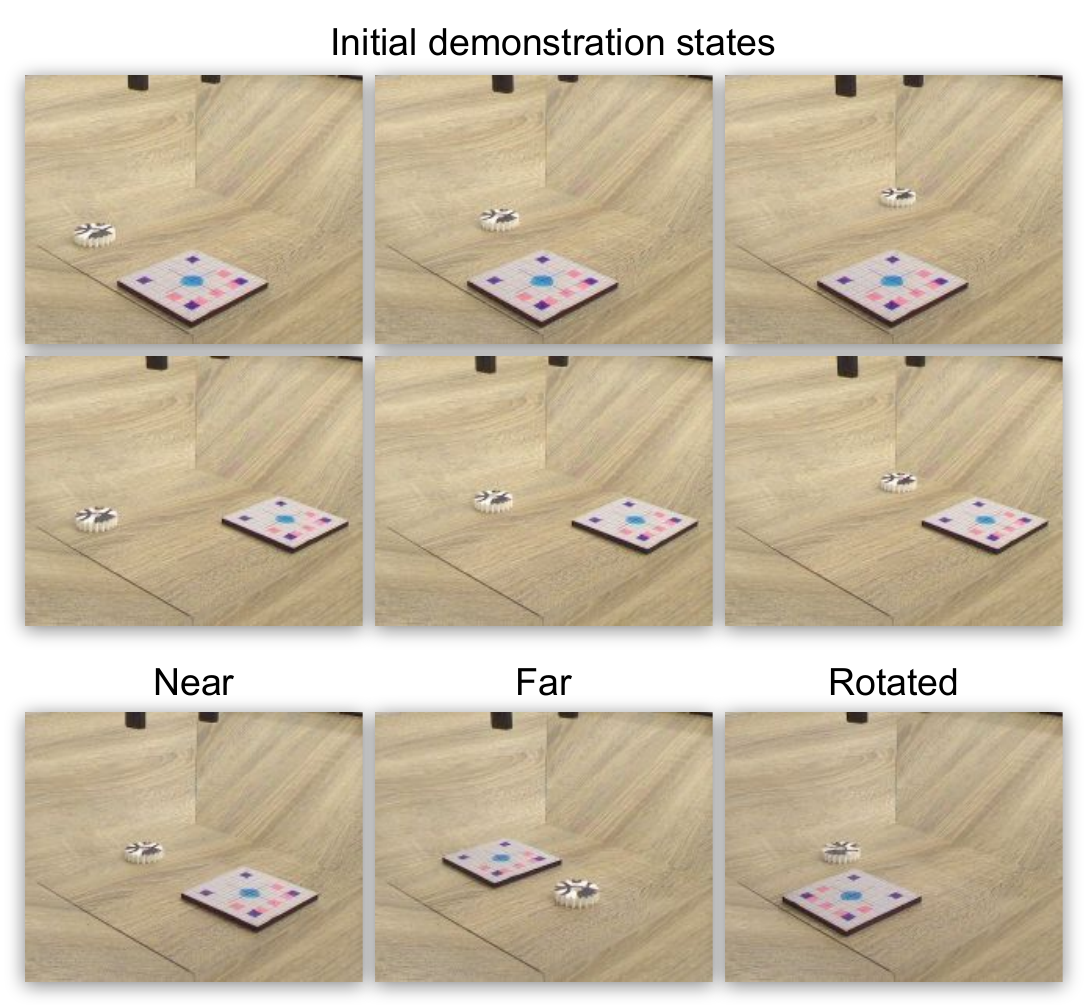}
  \caption{
\small Examples of initial states from the precision placement experiments.
The top 2 rows show the initial states which were used for the 6 demonstrations.
The bottom rows depicts examples of initial states from the evaluation sets.
Note that none of the evaluation states have been seen within the demonstrations.
During the Near setting we placed the target to a location between the demonstrated locations.
For the Far setting we used an initial state where the gear was on the other side of the target.
Finally in the Rotated setting we moved even further from the demonstration distribution where we rotated the target by 90 degrees and placed it in a previously unseen location. 
}
  \label{fig:precision_examples}
  \vspace{-0.2cm}
\end{figure}
\begin{table}[t]
\resizebox{\linewidth}{!}{%
\begin{tabular}{l|cc|cc}
\toprule
 & $x/mm$ & $\sqrt{var(x)}/mm$ & $y/mm$ & $\sqrt{var(y)}/mm$ \\
\midrule
Demos & 0.96 & 0.63 & -2.29 & 1.42 \\
Near & -4.66 & 0.63 & -3.82 & 2.00 \\
Far & -5.57 & 1.03 & -4.14 & 1.64 \\
Rotated & -1.48 & 1.27 & -2.85 & 1.15 \\
\bottomrule
\end{tabular}%
}
\caption{\small Evaluations of precision of object placement for our controller. This table contains the horizontal and vertical means and spreads seen across the demonstrations as well as across the 3 evaluation positions.
For demonstrations this was computed from the 6 demonstrations and for each target these values are calculated from 30 trials.}
\label{tab:precision_table}
\end{table}

\section{Extracting motion from demonstrations}
\label{sec:appendix_hypers}

In section \sref{sec:decomosition} we described how we split the motion into sections and how we extract active points.
Since this is a part where we did task specific parameter selection we would like to describe the process in greater detail here. 
Assuming we already have clustering, this process has 2 main stages: 1) temporal segmentation and alignment or demonstrations 2) active point extraction.

\textbf{Temporal segmentation.}
The core idea is that we noticed that for all the tasks we considered we can segment the motion based on the gripper actuation or external forces.
We use these signals to construct a motion plan which is then executed by the controllers.
To extract gripper actuation events we consider the gripper openness positions and note points where the position crosses a selected threshold.
These time points are beginnings or ends of grasps.
When manipulating small objects it was helpful to not open the gripper fully as that gave a higher grasp precision to the demonstrator and therefore we used a different threshold in each task.
This threshold is noted in \tref{sec:appendix_hypers} as \textit{Gripper open}.
The second type of an event we consider is a beginning or an end of a force phase.
To extract these we look at the vertical force measured by our force torque sensor and first smoothen the signal with a kernel with variance of 2.5 seconds.
Then we define a local force maximum by computing $f_{local-max}$ within a 5s windows.
We then construct a normalized force signal as $f_{norm} = f / max(f_{threshold}, f_{local-max})$ where $f_{threshold}$ is noted in \tref{sec:appendix_hypers} as \textit{Max force}.
This ensures that our normalized forces are between 0 and 1, but the normalized forces stay small when no forces are being applied.
We define a force event when the normalized force crosses a threshold of 0.5 as these will be points where we either engage in force-feedback behaviour or disengage from it.

When executing a force feedback behaviour we modify the visual servoing controller.
Instead of following the vertical action from the visual servoing we apply a P controller on vertical component to keep a force of $1.5N$.
We also disallow advancing to the next frame unless the current force is within $\pm 0.3N$ of this target.

In order to compare information across demonstrations we resample each segment of each demonstration into a fixed lenght block.
However resampling linearly across time leads to oversampling of static phases.
Therefore we resample the trajectories along the distance traveled.
This is done using a per-timestep motion metric $t_m = ||v_{end-effector}||^2 / 0.04 + ||v_{fingers}||^2 / 0.5$ where $v_{end-effector}$ is the linear velocity of the end-effector in $m/s$ and $v_{fingers}$ captures the opening and closing of the gripper.
$t_m$ is then used to linearly interpolation within each motion segment to align all of the demonstrations.

The final step is to extract constant motion primitives which are needed because, after a grasp or a release of an object the gripper occludes large potion of the workspace.
To do this we look at the standard deviation across the demonstrations at every time point $v^t_{std}$, the minimum vertical velocity $v^t_{min}$ and the maximum velocity reached during the demonstration $v_{max}$.
If at any timepoint all demonstrations reliably contain the motion (\ie $v^t_{min} > 0.5 v^t_{std}$) and the motion is significant (\ie $v^t_{min} > 0.03 v_{max}$) we consider this a reliable signal that all demonstration contain the same motion.
\textbf{Active point extraction.}
As discussed in \sref{sec:decomosition} we extract active points in several steps: We do a first pass in selecting active point candidates, cluster points and use points to vote for a clusters, and clean up the points within the selected clusters.

To do a first pass in selecting active points we use 3 seemingly trivial, but powerful heuristics: a) Reject points which are not currently visible b) Reject points which do not move c) Reject points which end up at different image locations.
For a) we look at the final frame of the motion and define a threshold called \textit{Saliency} in \tref{sec:appendix_hypers} for the fraction of demonstrations where the point must be visible in.
Then for b) we reject points whose overall motion during the segment is less than a \textit{Is gripper} fraction of the 90th quantile across the points.
For c) we compute the variance of the final point location across demonstrations and require that it is less than a threshold.
We called the threshold for the square root of the variance the \textit{Cross-demo variance}.

After this procedure we have a set of points which belong the important object for a given segment.
In many cases these points would be good enough to use as active points for the final visual servoing.
One of the limitations of TAPIR, as most point tracking models, however is its ability to track points across very large changes of scale which is something we observe in our tasks as we use the gripper camera.
Points chosen above are descriptive of the final frame which is often up-close and therefore those features don't tend to be detected when the gripper is further out or when it is looking at the object from a different direction.
To get points representative of the whole motion we leverage the clustering.
We first clean the clusters by removing points which could not be assigned to any cluster with an error of less than \textit{Average Error removal}.
Then we count how many of the points belong to which cluster and normalize the counts by dividing by the largest number of votes.
Then we merge all clusters which have score over a given threshold called \textit{Multi cluster}.
At this point we have a bank of relevant points which will work across all scales.

Lastly this clustering leads to a potentially large number of points.
To remove them we compute an average visibility score across the whole motion and remove all points below this score.
This threshold is called \textit{2nd pass saliency} in \tref{sec:appendix_hypers}.
This last filtering step removes outliers leftover from the clustering process.
To remove these we for each point computed the distance to the \textit{closest K} point and averaged this across the frames where it was visible.
Then we removed points where this was above a given threshold \textit{Dist}.
If at this point we had more than 128 points, we randomly selected which ones are going to be used during execution of the motion.

We believe that many of the hyper-parameters in \tref{sec:appendix_hypers} could be easily automatically extracted without the need to be manually specified by a human.
However, in our case it was very easy to select and tune these by looking at visualisations from the intermediate steps of the process.
For most of them we selected the value by looking at the distribution of the variable it was thresholding.
Therefore we have not attempted to automate this process further at this stage.

\begin{table*}
\centering
{\small
\begin{tabular}{|l|l|l|l|l|l|l|l|l|l|l|}
\toprule
 & \parbox{1cm}{Gripper\\Open} & \parbox{1cm}{Max\\Force} & \parbox{1cm}{Saliency} & \parbox{1cm}{Is\\Gripper} & \parbox{1cm}{Cross Demo\\Var.} & \parbox{1cm}{Avg Err\\Removal} & \parbox{1cm}{Multi\\Cluster} & \parbox{1cm}{2nd Pass\\Saliency} & \parbox{0.7cm}{Closest K} & \parbox{1cm}{Dist.} \\
\midrule
Apple On Jello & 0.04 & 8 & 0.5 & 0.5 & 30 & 0.01 & 0.5 & 0.2 & 4 & 20 \\
Juggling Stack & 0.13 & 10 & 0.4 & 0.3 & 20 & 0.0005 & 0.8 & 0.5 & 5 & 30 \\
Four object puzzle & 0.41 & 40 & 0.5 & 0.5 & 30 & 0.01 & 0.5 & 0.7 & 10 & 20 \\
Lego Stack & 0.15 & 40 & 0.5 & 0.5 & 30 & 0.01 & 0.5 & 0.9 & 3 & 30 \\
Tapir to robot V2 & 0.15 & 40 & 0.5 & 0.5 & 30 & 0.01 & 0.5 & 0.9 & 3 & 30 \\
Gluing & 0.13 & 10 & 0.4 & 0.3 & 20 & 0.0005 & 0.8 & 0.5 & 5 & 30 \\
Gear on grid & 0.41 & 40 & 0.5 & 0.5 & 30 & 0.01 & 0.5 & 0.7 & 10 & 20 \\
Pass Butter & 0.15 & 10 & 0.8 & 0.1 & 60 & 0.01 & 0.5 & 0.5 & 5 & 30 \\
\bottomrule
\end{tabular}
\caption{Summary of all of the parameters tuned while processing the demonstrations.}
}
\end{table*}

\section{Controller Implementation Details}
\label{sec:controller_details}

Given a set of target points $g_t$ and corresponding detected points $p_t$, we aim to compute an action that would minimize the error, using a linear approximation of the function mapping actions to changes in $p_t$.  
We then take a step in that direction and repeat. This process can be summarized as:
\begin{equation}
    v_{vs} = argmin_{v_t} ||J_{p_t} v_t - (g_t - p_t)||^2
\label{eq:simple_velocity_controller}
\end{equation}
However, in practice this approximation has two main issues: the optimal set of detected and target points to use for servoing may change depending on the current state, and the Jacobian of the error may not be a good approximation of the desired motion.  These could likely be rectified by learning, but for simplicity, in this paper we use simple heuristics to resolve them analytically in ways that work for (approximately) rigid objects.  

\textbf{Point selection:}
TAPIR provides a visibility score $vis$ between 0 and 1.
We use this score by summing the visibility across the demonstration frame which is being followed and the current frame and only using the 30\% most visible or confidently-detected points.

\textbf{Target selection:}
Recall that motion plans consist of a series of temporal segments of the demonstrations, along with a set of \textit{active points} for each segment.  Our low-level motion planner requires a \textit{single} servoing target for every point at every timestep, but we have a full set of trajectories for each demonstration.  How can we summarize this to a single point?  In order to have precision, we would like to average across trajectories to reduce noise from the point tracking.  However, simple averaging across the whole trajectory may not be sensible, since the initial configurations may be diverse.  For many tasks (e.g. gluing), we must follow the entire trajectory.  Thus, we take a hybrid approach: at the beginning of each temporal segment, we choose a single demonstration (the nearest one) and follow the full trajectory until the final frame of the demos, at which point we average across all available demos.  This is a reliable procedure because we assume that our placement tasks form a `funnel'--i.e., the demonstrations may start off with objects in diverse states, but the object of interest always converges to the same location.

Specifically, we split every visual servoing motion into 2 phases: 1) demonstration follow phase and 2) the final phase.
Phase (1) allows us to repeat the broad motion while phase (2) allows us to reach a consistent final state.
At the beginning of every visual servoing phase we pick a demonstration to follow: we measure the Euclidean distance between the visible points on initial frame of every demonstration and the currently-detected locations, and pick the one with the lowest distance.
We servo toward the initial location until the 30th percentile of per-point errors falls below a threshold, at which point we advance to the next frame of the same demonstration.  
We alternately servo when the error is above the threshold and advance when the error is below the threshold, until we reach the final frame of the demonstration, at which point the final phase begins.
For the final phase (2), we want to achieve a consistency of behaviour no matter which demonstration was followed.
We use the (visibility-weighted) average of points across demonstration as the target $g$.
To move on to the next timestep in phase 1 we use a threshold of 12 pixels and phase 2, we use a threshold of 2 pixels, but we multiply this threshold by 1.01 for every timestep we spend in this phase, thus gradually increasing the termination threshold to ensure that the temporal segment is guaranteed to terminate even in the presence of noise.
\textbf{Computing the Jacobian}
Given a set of target points, we must next compute a Jacobian with respect to the detected points, which will allow us to compute an action which will minimize the error.  However, when the distance between the target points and the detected points is large, the linear approximation implied by the Jacobian may be poor: in particular, the model may use rotation and z-motion to explain errors due to in-plane translation.  For example, when the object is far from the image center, the rotation and translation are highly correlated in the motions they will induce in the target points.  Therefore, we use Gram-Schmidt to orthogonalize the last 4 columns of the Jacobian relative to the first two.  Recall that our Jacobian can be written as: 
\begin{equation}
J = 
\begin{bmatrix}
1 & 0 & -u & -uv & 1 + u^2 & -v \\ 
0 & 1 & -v & -1 - u^2 & uv & u 
\end{bmatrix}
\label{eq:jacobian_full}
\end{equation}
Where $u$ is the x-position of a detected point, and $v$ is the y-position.  Thus, the first two columns correspond to $x$ and $y$ translation, the third is $z$ translation, fourth and fifth are $x$ and $y$ tilt, and sixth is rotation.
$x$ and $y$ translation dominates the loss, and the other columns are highly correlated with the first two.
Columns 4 and 5 describing tilt are only used in the simulated ablations described in \sref{sec:simulated_ablations}.
We orthogonalize the last 4 columns of the Jacobian with respect to the first two, after evaluating it on our points using the Gram–Schmidt Orthogonalisation method and denote it $J^{\perp}_{p}$.
This leads to the final controller equation for the visual servoing command $v_{vs}$:
\begin{equation}
\begin{split}
    v_{vs} = &\frac{1}{2} ( argmin_{v_g} ||J^{\perp}_{p} v_g - (g - p)||^2 \\ 
    & - argmin_{v_g} ||J^{\perp}_{g} v_g - (p - g)||^2
    )
\end{split}
\label{eq:velocity_controller}
\end{equation}

\textbf{Statistical Bias and the Jacobian}  
As described above, a key problem with the Jacobian is statistical bias in which noise in the (2D) detections increases the estimate of action components which decrease the scale of the points (\eg -z for a wrist-camera).  
We therefore modify the Jacobian so that it performs variance scaling for z-translation.  Specifically, 
We flip the role of $p$ and $g$ in the solver, i.e. we compute both the standard Jacobian update $v_{vs}$, and also the \textit{inverse} Jacobian update $\bar{v}_{vs}$ which treats the demonstration points $g$ as controllable, and moves them toward the current states.  The controller update is then $(v_{vs}-\bar{v}_{vs})/2$.  It is straightforward to show that this does not modify the $v_x$ and $v_y$ component of the motion, as the motion and inverse in-plane rotation are identical.  Therefore, in this section, we focus on showing that the update for the $v_z$ component becomes variance matching.  To show this, we focus on only the Jacobian with respect to the $z$ component (recall that $x$ and $y$ are orthogonalized out from this).  Thus, we have:

$$argmin_{v_z} ||J_{z} v_z - (g - p)||^2$$

The $z$-component of the Jacobian $J^z_p$ is simply $[-u_p -v_p]^{\top}$, the $-x$ and $-y$ components of $p$.
The solution to this equation is $J^{z-1}_{p} \cdot (p - g)$, where $J^{z-1}_{p}$ is the pseudinverse of $J^{z}_p$, which can be computed analytically as $-p / ||p||^2$.
In the presence of noise, we have that $p = (\hat{p}+\epsilon_{p})$, where $\hat{p}=[u_c,  v_c]^{\top}$ is the true point locations, and $\epsilon_{p}$ is the noise with variance $\sigma_p^2$ introduced by the detector in the current detections, which we assume is zero mean and uncorrelated with any other variables; we can define an analogous decomposition for the target points.  Substituting and taking the expectation with respect to the noise yields an update of the form:

\begin{equation*}
    v_z = \frac{-\hat{p} - \epsilon_{p}}{||\hat{p} + \epsilon_{p}||^2} (\hat{g} + \epsilon_{g} - \hat{p} - \epsilon_{p})
\end{equation*}

Because the $\epsilon$ values are uncorrelated with other terms, if we consider the expected value, this simplifies to:

\begin{equation*}
    \mathbb{E}[v_z] = \frac{||\hat{p}||^2 - \hat{p} \cdot \hat{g} +\sigma_p^2}{||\hat{p}||^2+\sigma_p^2}
    = 1 - \frac{\hat{p} \cdot \hat{g}}{||\hat{p}||^2+\sigma_p^2}
\end{equation*}

From this equation we can see that even if we reached the goal state, \ie $\hat{p} = \hat{g}$, our controller would be outputting a $v_z > 0$.
Specifically it would be outputting an action of $1 / (1 + g^2 / \sigma_p^2)$ coming from the zoom-out bias.
Computing the analogous term from a Jacobian in the other direction and subtracting it yields:

\begin{equation*}
\begin{split}
    \mathbb{E}[v_z - \bar{v}_z] &= - \frac{\hat{p} \cdot \hat{g}}{||\hat{p}||^2+\sigma_p^2} + \frac{\hat{g} \cdot \hat{p}}{||\hat{g}||^2+\sigma_g^2} \\
    &= \hat{p} \cdot \hat{g} \left (\frac{1}{||g||^2} - \frac{1}{||p||^2} \right)
\end{split}
\end{equation*}

We can see that this expression is zero when $||g||^{2}=||p||^{2}$ which can be interpreted as matching variances.
\section{RoboTAP Dataset}

\textbf{Video sources} The videos come from DeepMind Robotics real-world data collections from Manipulation Task Suite (MTS). To make the data source diverse, we randomly sample a subset from the publicly released Sketchy dataset~\cite{cabi2019scaling}, S3K dataset~\cite{vecerik2020s3k} and NIST dataset~\cite{bousmalis2023robocat}. The Sketchy data are sensed and recorded with 3 cage cameras and 3 wrist cameras (wide angle and depth), focusing on 3 variable-shape rigid objects coloured red, green and blue (rgb dataset) and 3 deformable objects: a soft ball, a rope and a cloth (deformable dataset). The S3K data contains a collection of rigid and nonrigid objects: \eg shoes, toys. The NIST data contains gears and board that is widely known as the NIST challenge in Robotic Grasping and Manipulation~\cite{sun2021research}. Figure~\ref{fig:dataset} visualize the examples of RoboTAP videos and the annotated point tracks.

\begin{figure*}[t]
\centering
\begin{tabular}{ccc}
    \includegraphics[height=0.15\linewidth]{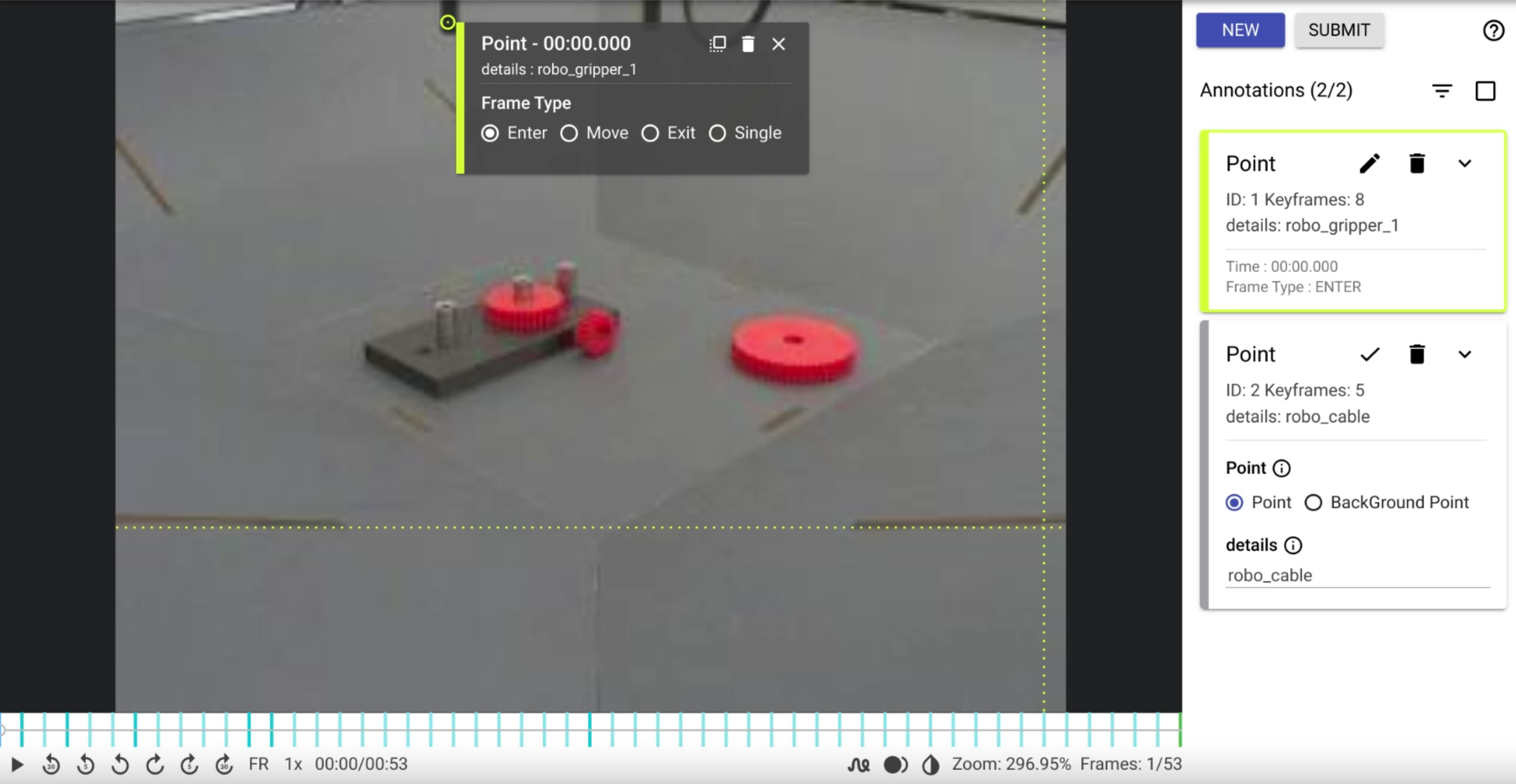} &
    \includegraphics[height=0.15\linewidth]{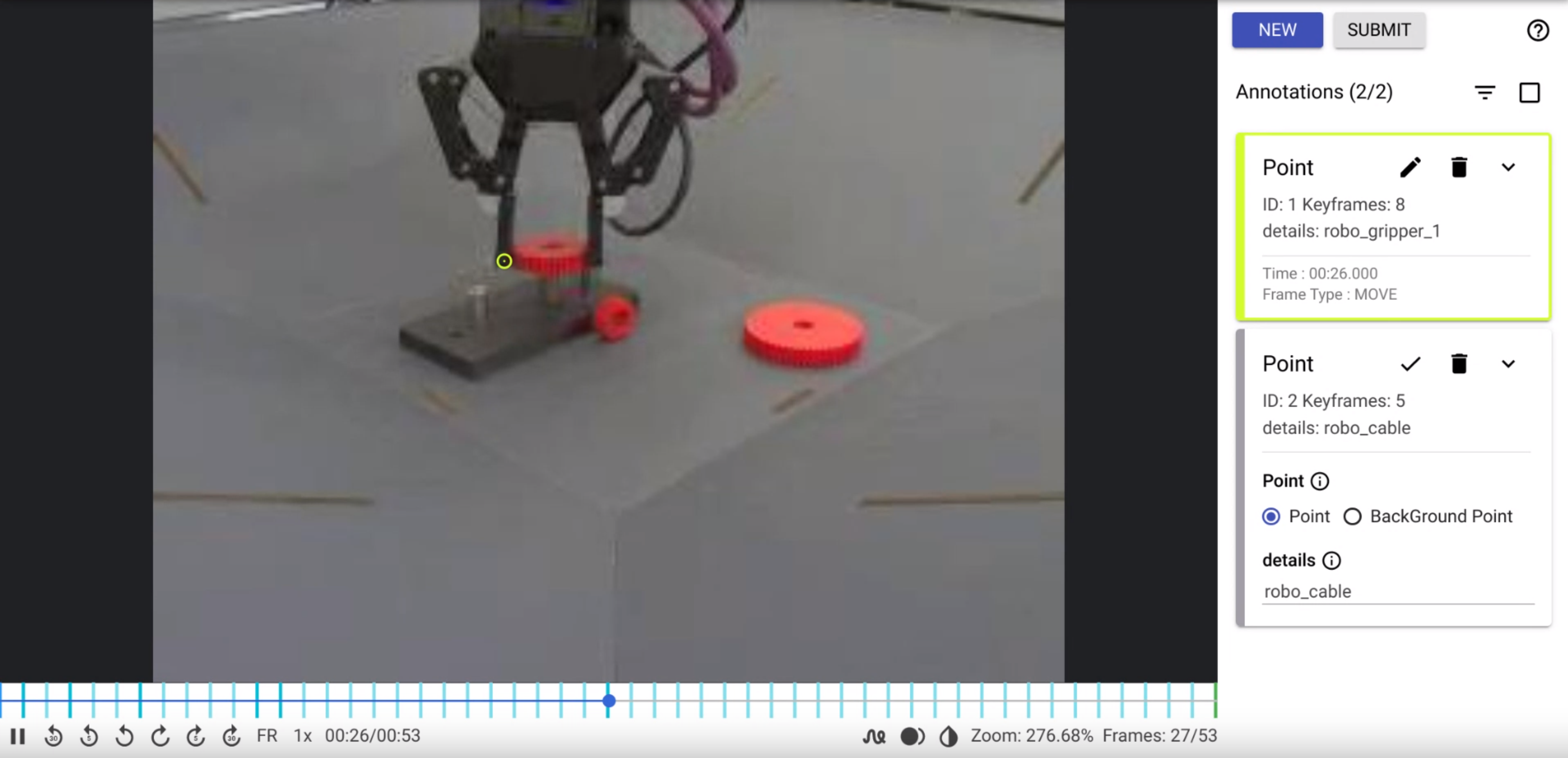} &
    \includegraphics[height=0.15\linewidth]{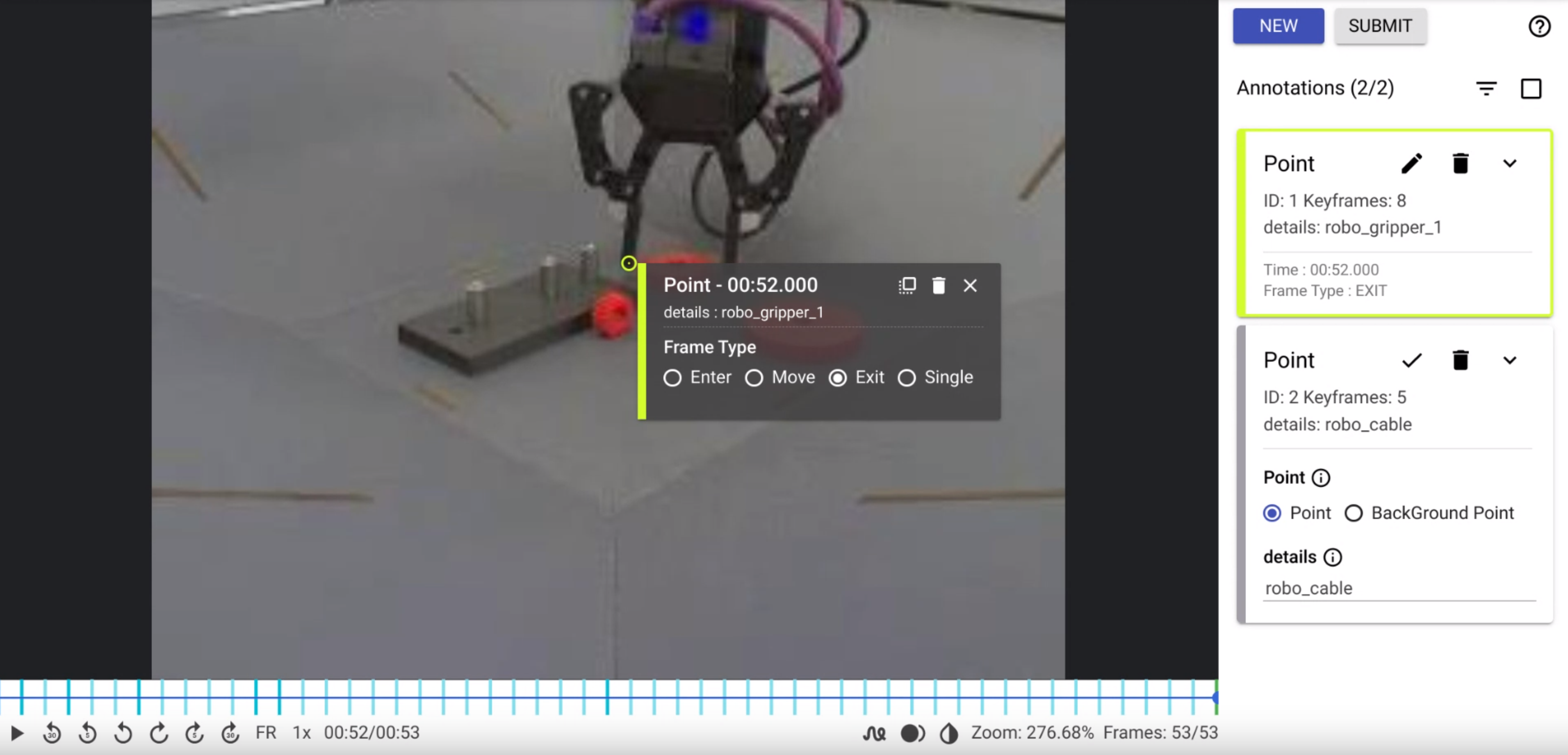} \\
    Beginning Frame & Middle Frame & End Frame \\
\end{tabular}
\caption{Illustration of annotation interface and process for RoboTAP dataset. The interface consists of three components: visualization panel, buttons, and information panels. Each track begins with an \texttt{ENTER} point, continues with \texttt{MOVE} points, and finishes when the annotator sets it as an \texttt{EXIT} point. The optical flow track assist algorithm~\cite{doersch2022tap} is used to interpolate the intermediate trajectories to improve annotation accuracy and efficiency.}
\label{fig:annotation}
\end{figure*}
\begin{figure*}[t]
\centering
\begin{tabular}{ccc}
    \includegraphics[height=0.17\linewidth]{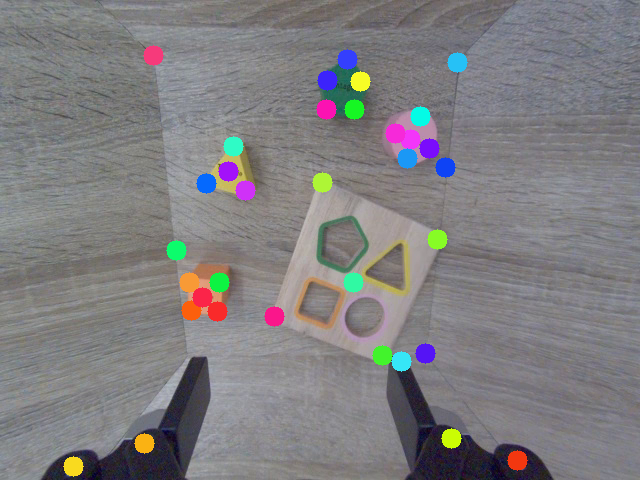} &
    \includegraphics[height=0.17\linewidth]{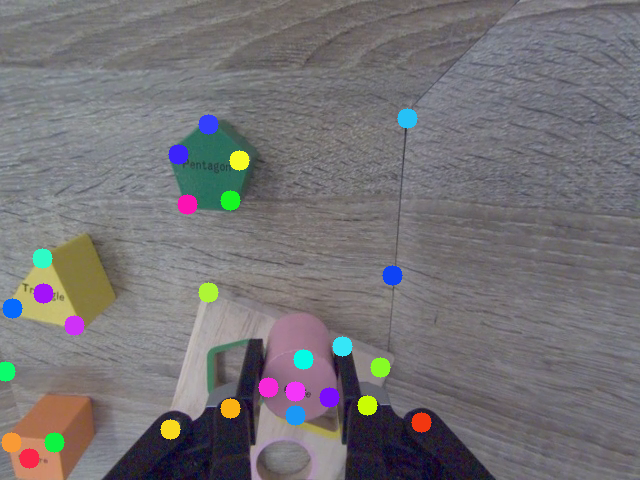} &
    \includegraphics[height=0.17\linewidth]{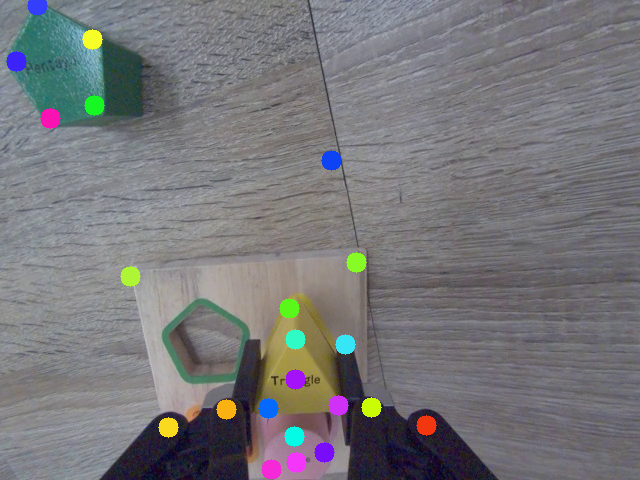} \\
    \includegraphics[height=0.17\linewidth]{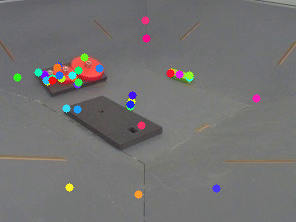} &
    \includegraphics[height=0.17\linewidth]{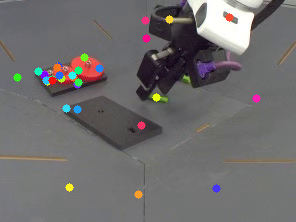} &
    \includegraphics[height=0.17\linewidth]{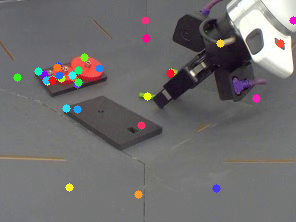} \\
    \includegraphics[height=0.17\linewidth]{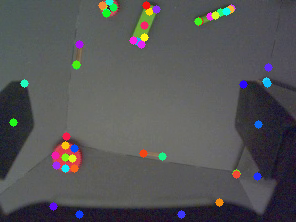} &
    \includegraphics[height=0.17\linewidth]{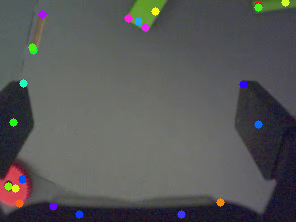} &
    \includegraphics[height=0.17\linewidth]{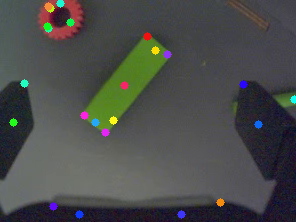} \\
    \includegraphics[height=0.17\linewidth]{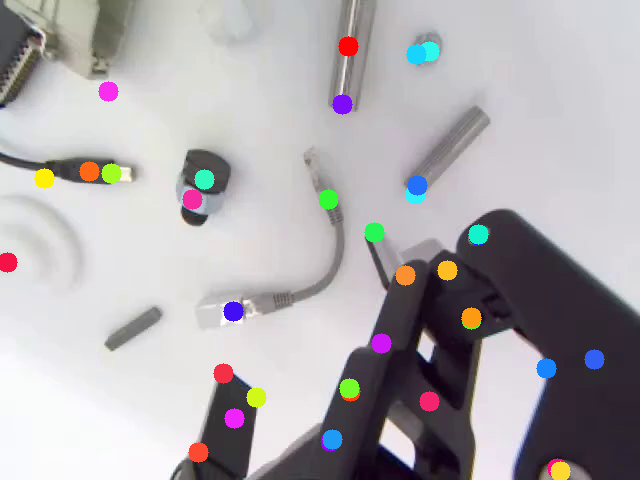} &
    \includegraphics[height=0.17\linewidth]{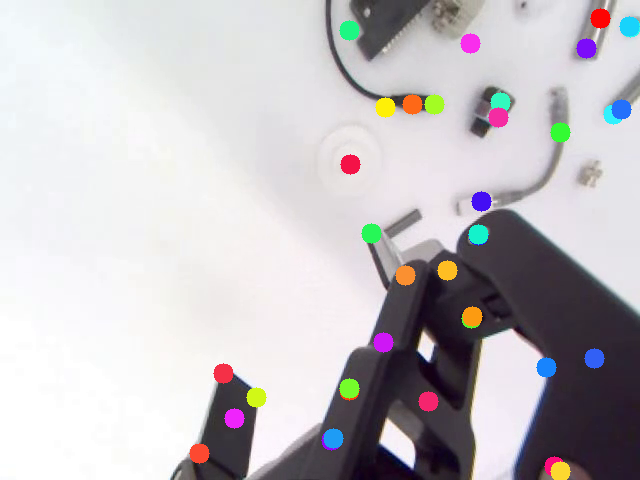} &
    \includegraphics[height=0.17\linewidth]{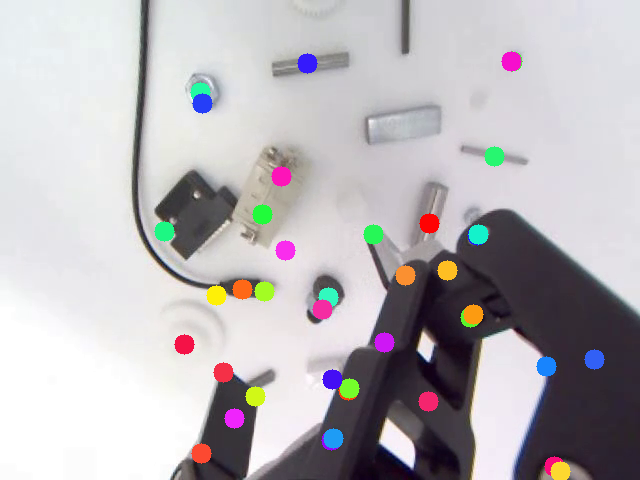} \\
    \includegraphics[height=0.17\linewidth]{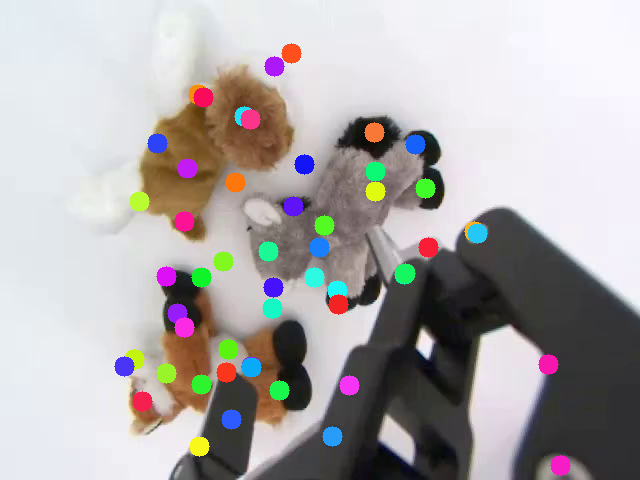} &
    \includegraphics[height=0.17\linewidth]{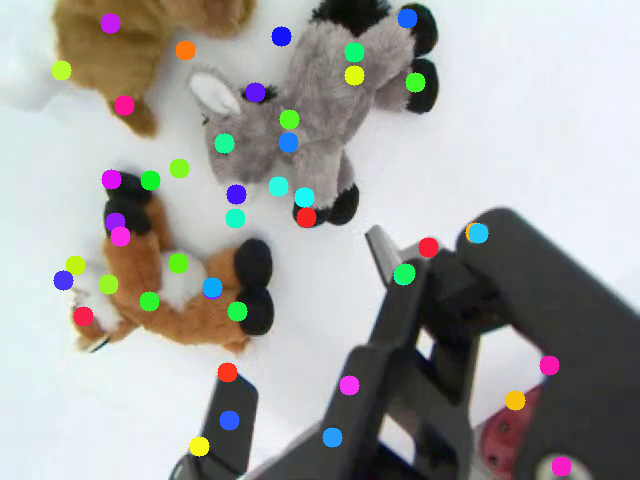} &
    \includegraphics[height=0.17\linewidth]{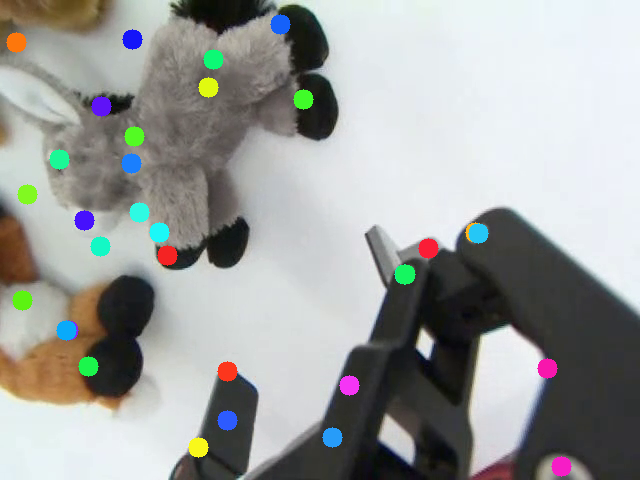} \\
    \includegraphics[height=0.17\linewidth]{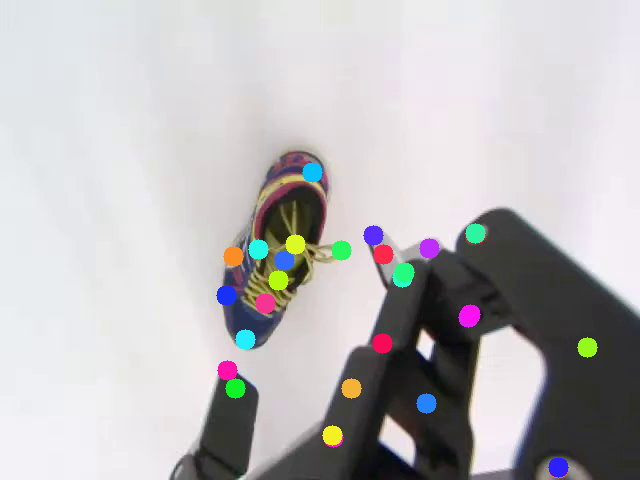} &
    \includegraphics[height=0.17\linewidth]{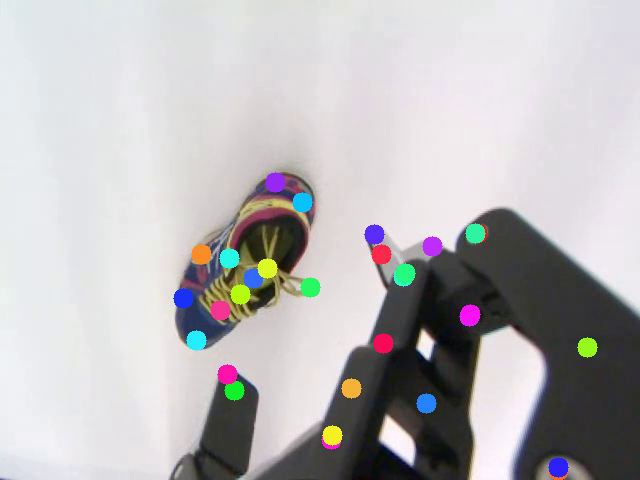} &
    \includegraphics[height=0.17\linewidth]{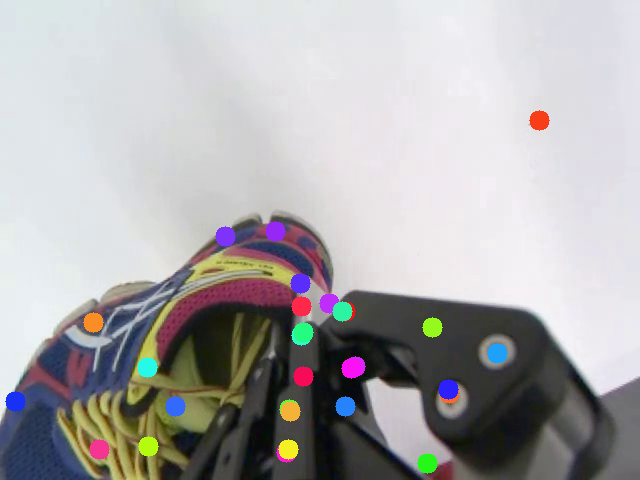} \\
    \includegraphics[height=0.17\linewidth]{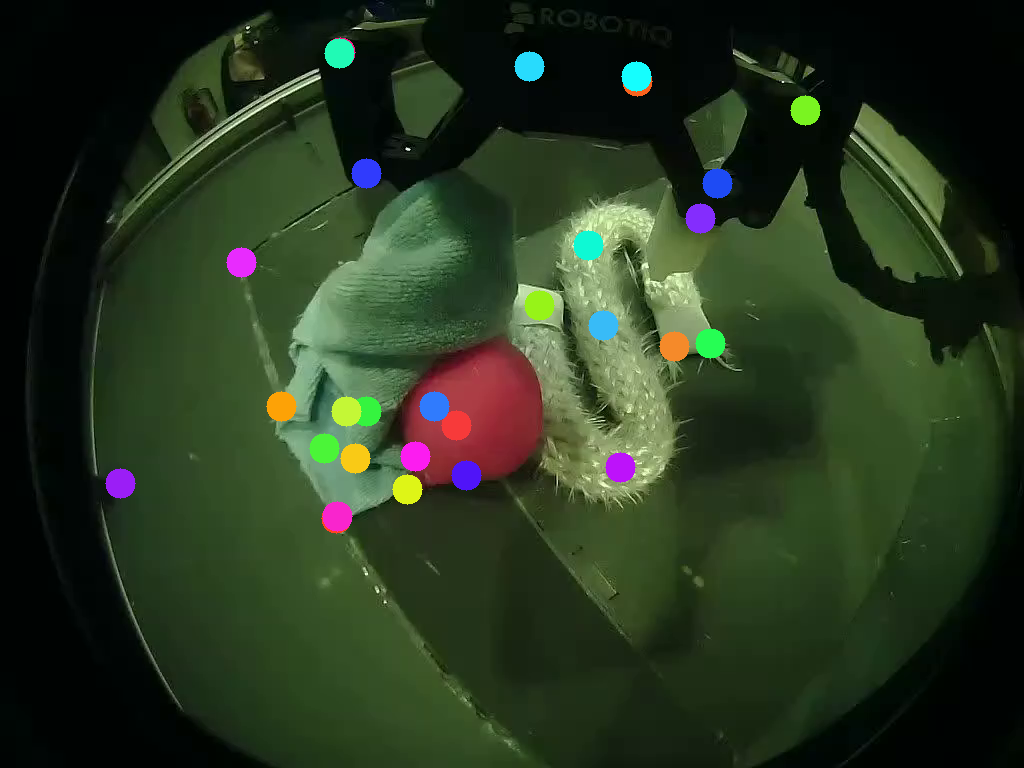} &
    \includegraphics[height=0.17\linewidth]{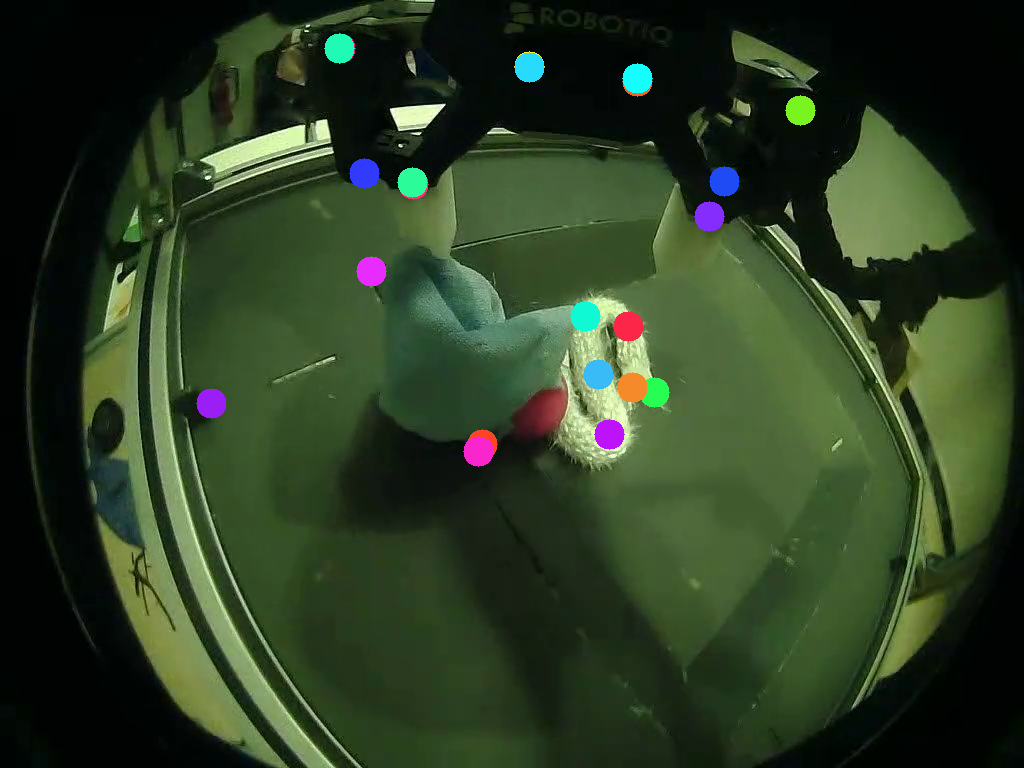} &
    \includegraphics[height=0.17\linewidth]{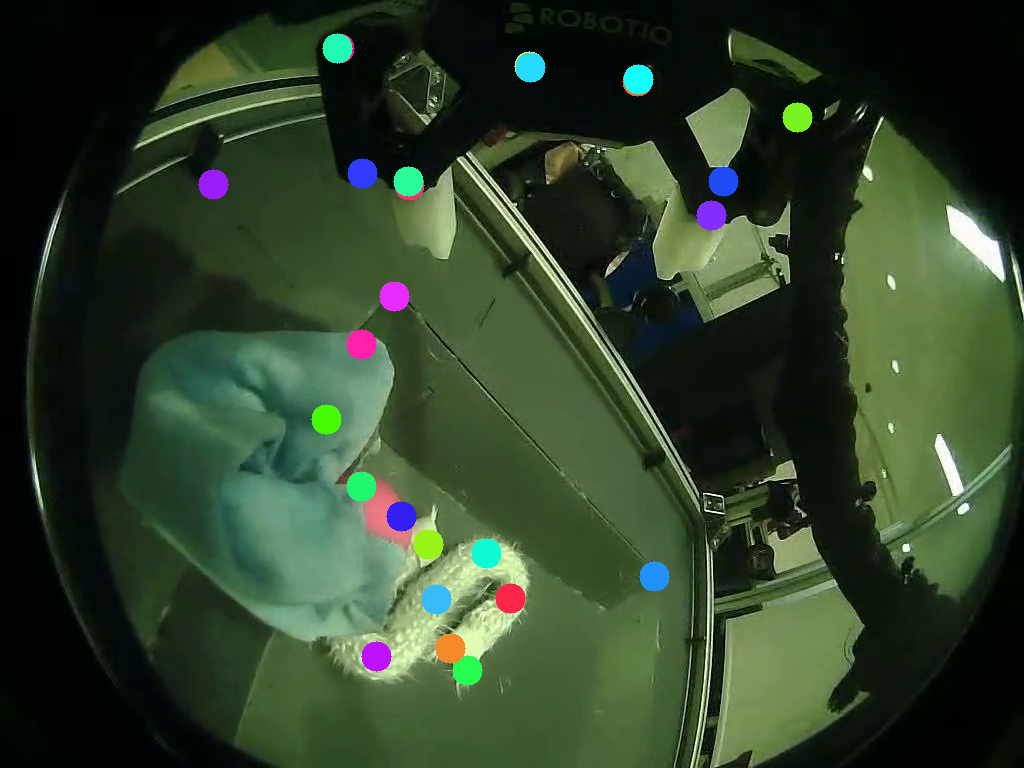} \\
\end{tabular}
\caption{Examples from RoboTAP: we show a few examples annotations with 3 frames per video.}
\label{fig:dataset}
\end{figure*}

\textbf{Annotation interface} Figure~\ref{fig:annotation} illustrates the annotation interface. The point annotation interface presented to annotators is based on~\cite{kuznetsova2021efficient}. The interface loads and visualizes video in the visualization panel. Buttons in the video play panel allow annotators to navigate frames during annotation. The information panel provides basic information, e.g., the current and total number of frames. The annotation buttons (\texttt{NEW} and \texttt{SUBMIT}) allow annotators to add new point tracks or submit the labels if finished. The label info panel shows each annotated point track and the associated `tag' string. The cursor is cross shaped which allows annotators to localize more precisely.
\textbf{Annotation process} For acquisition of high-quality and diverse point tracks dataset, we encourage the annotators to choose any point that can be tracked reliably. Each track begins with an \texttt{ENTER} point, continues with \texttt{MOVE} points, and finishes when the annotator sets it as an \texttt{EXIT} point. Annotators can restart the track after an occlusion by adding another \texttt{ENTER} point and continuing. An optical flow based track assist algorithm is also utilized to interpolate the point tracks in the intermediate frames, this helps improve the annotation stability, accuracy and productivity.  More details are provided in ~\cite{doersch2022tap}.

\textbf{Dataset statistics} Figure~\ref{fig:stats} shows more detailed statistics of RoboTAP dataset based on histograms. There are mainly 3 different video resolutions: 222 x 292, 480 x 640 and 768 x 1024. Although the total number of videos is not large, we find the point tracks cover a quite broad range of interests. For example, the majority of contiguous point tracks spans between 20 frames to 150 frames, with some longer than 1000 frames (\eg 40 seconds). Among 11592 point tracks, around 5000 point tracks are visible across the episode, with the rest being occluded in a relatively uniform distribution. The most extremely moving points can reach a maximum distance that covers the whole video (i.e. from left to right) while a majority moves in a quarter of the image size.

\begin{figure*}[t]
\centering
\begin{tabular}{cc}
    \includegraphics[width=0.45\linewidth]{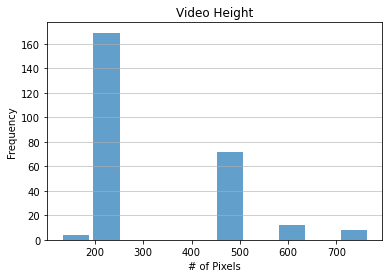} &
    \includegraphics[width=0.45\linewidth]{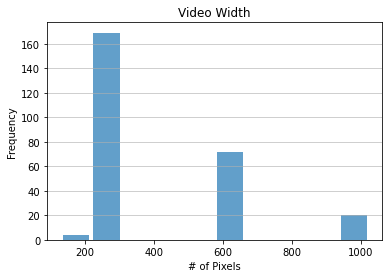} \\
    \includegraphics[width=0.45\linewidth]{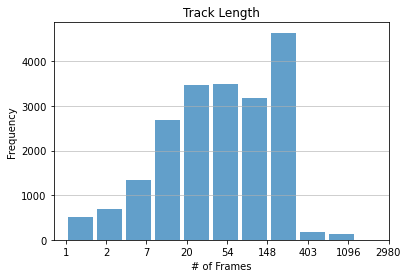} &
    \includegraphics[width=0.45\linewidth]{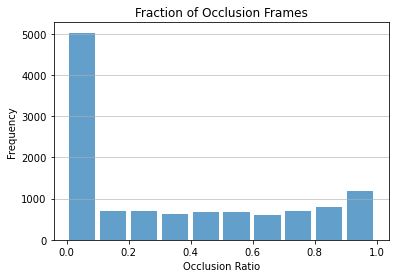} \\
    \includegraphics[width=0.45\linewidth]{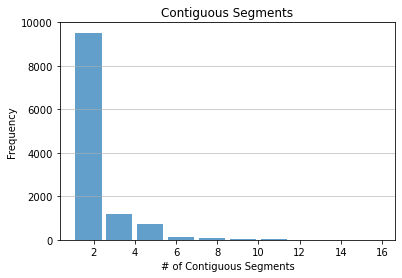} &
    \includegraphics[width=0.45\linewidth]{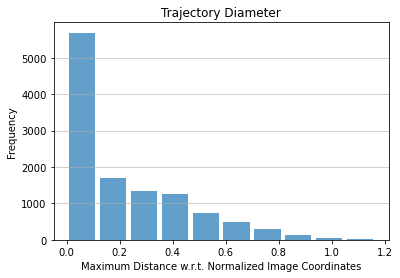} \\
\end{tabular}
\caption{Statistics of trajectories in RoboTAP dataset. \emph{Video Height} and \emph{Video Width} refers to the original resolution of the videos. \emph{Trajectory Length} refers to the number of frames a contiguous point tracks contains. \emph{Fraction of Occluded Frames} refers to the proportion that a point is occluded in the video. \emph{Contiguous Segments} refer to the different number of contiguous sections of point trajectories with breaks due to occlusion. \emph{Trajectory Diameter} refers to the maximum distance between the positions of a point over time.}
\label{fig:stats}
\end{figure*}

\fi

\end{document}